\documentclass[11pt]{article}

\usepackage[hyphens]{url}
\usepackage{microtype}
\usepackage{graphicx}
\usepackage{booktabs}
\usepackage{amsmath}
\usepackage{amsfonts}

\usepackage{amssymb}
\usepackage{bm}
\usepackage{enumitem}
\usepackage{xspace}
\usepackage{makecell}
\usepackage{multirow}
\usepackage{algorithm}
\usepackage{algorithmic}
\usepackage{wrapfig}
\usepackage{ziplab-tech-report}
\usepackage{hyperref}

\newcommand{\reporttitle}{Block3D: Efficient Text-to-3D Generation via Block-Wise Diffusion}
\newcommand{\reportshorttitle}{Block3D}
\newcommand{\reportauthors}{%
  Bowen Cui\textsuperscript{1,$\dagger$}, Weijie Wang\textsuperscript{1,$\dagger$,*}, Zeyu Zhang\textsuperscript{1},
  Yefei He\textsuperscript{1}, Mingda Lin\textsuperscript{1},\\
  Haoyu Zhao\textsuperscript{1}, Yuanyu He\textsuperscript{1}, Donny Y. Chen\textsuperscript{2},
  Feng Chen\textsuperscript{3,*}, Bohan Zhuang\textsuperscript{1}}
\newcommand{\reportaffiliations}{%
  \textsuperscript{1}ZIP Lab, Zhejiang University \quad
  \textsuperscript{2}Monash University \quad
  \textsuperscript{3}University of Adelaide}
\newcommand{\reportdate}{August 25, 2026}

\newcommand{\reportcode}{}
\newcommand{\reportproject}{\url{https://alexandertsui.github.io/block3d/}}
\newcommand{\reportemail}{%
  Weijie Wang: \href{mailto:wangweijie@zju.edu.cn}{wangweijie@zju.edu.cn};\quad
  Feng Chen: \href{mailto:chenfeng1271@gmail.com}{chenfeng1271@gmail.com}}
\newcommand{\reportorcid}{}
\newcommand{\reportkeywords}{Text-to-3D generation, block-wise diffusion, 3D shape generation, efficient inference}
\newcommand{\reportfootnote}{%
  \textsuperscript{$\dagger$}Equal contribution.\quad
  \textsuperscript{*}Corresponding authors.}
\newcommand{\reportpdfauthors}{Bowen Cui, Weijie Wang, Zeyu Zhang, Yefei He, Mingda Lin, Haoyu Zhao, Yuanyu He, Donny Y. Chen, Feng Chen, Bohan Zhuang}
\newcommand{\reportlogoheight}{0.62cm}

\techreportlabel{ZIP Lab Technical Report}
\techreportshorttitle{\reportshorttitle}
\techreportsetlogoheight{\reportlogoheight}
\techreportdate{\reportdate}
\techreportconference{}
\ifdefempty{\reportcode}{}{\techreportcode{\reportcode}}
\ifdefempty{\reportproject}{}{\techreportproject{\reportproject}}
\ifdefempty{\reportemail}{}{\techreportemail{\reportemail}}
\ifdefempty{\reportorcid}{}{\techreportorcid{\reportorcid}}
\techreportkeywords{\reportkeywords}
\ifdefempty{\reportfootnote}{}{\techreportfootnote{\reportfootnote}}

\techreportlogos{%
  \techreportlogo{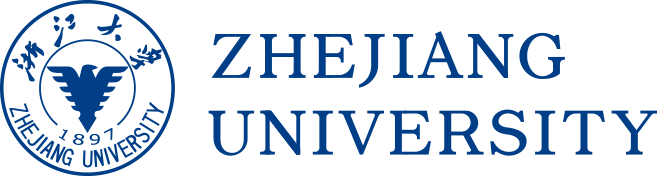}
  \techreportlogo{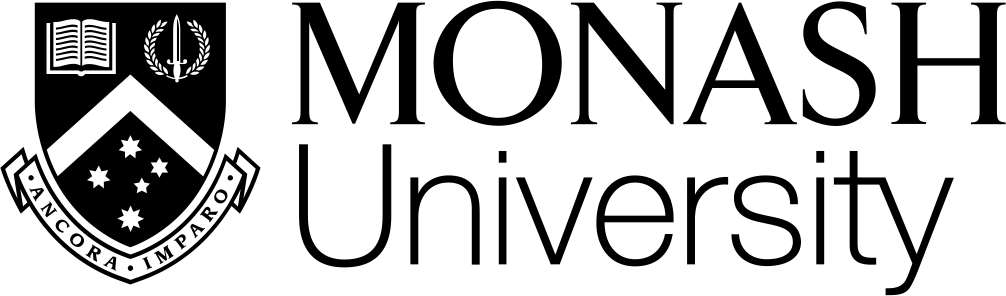}
  \techreportlogo{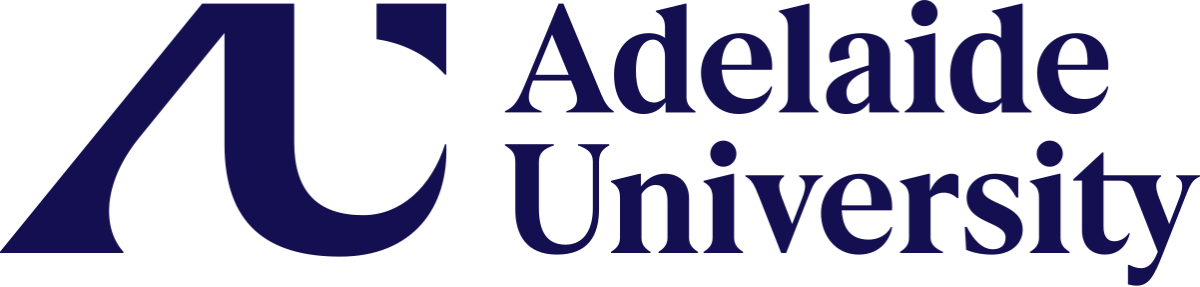}
}

\hypersetup{
  colorlinks=true,
  linkcolor=ziplabblue,
  citecolor=ziplabblue,
  urlcolor=ziplabblue,
  pdftitle={\reporttitle},
  pdfauthor={\reportpdfauthors},
  pdfsubject={ZIP Lab technical report},
  pdfkeywords={\reportkeywords}
}

\begin{document}

\ziplabmaketitle{\reporttitle}{\reportauthors}{\reportaffiliations}{%
  \begin{center}
  \begin{minipage}{\textwidth}
    \centering
    \vspace{0.04cm}
    \includegraphics[width=0.94\textwidth]{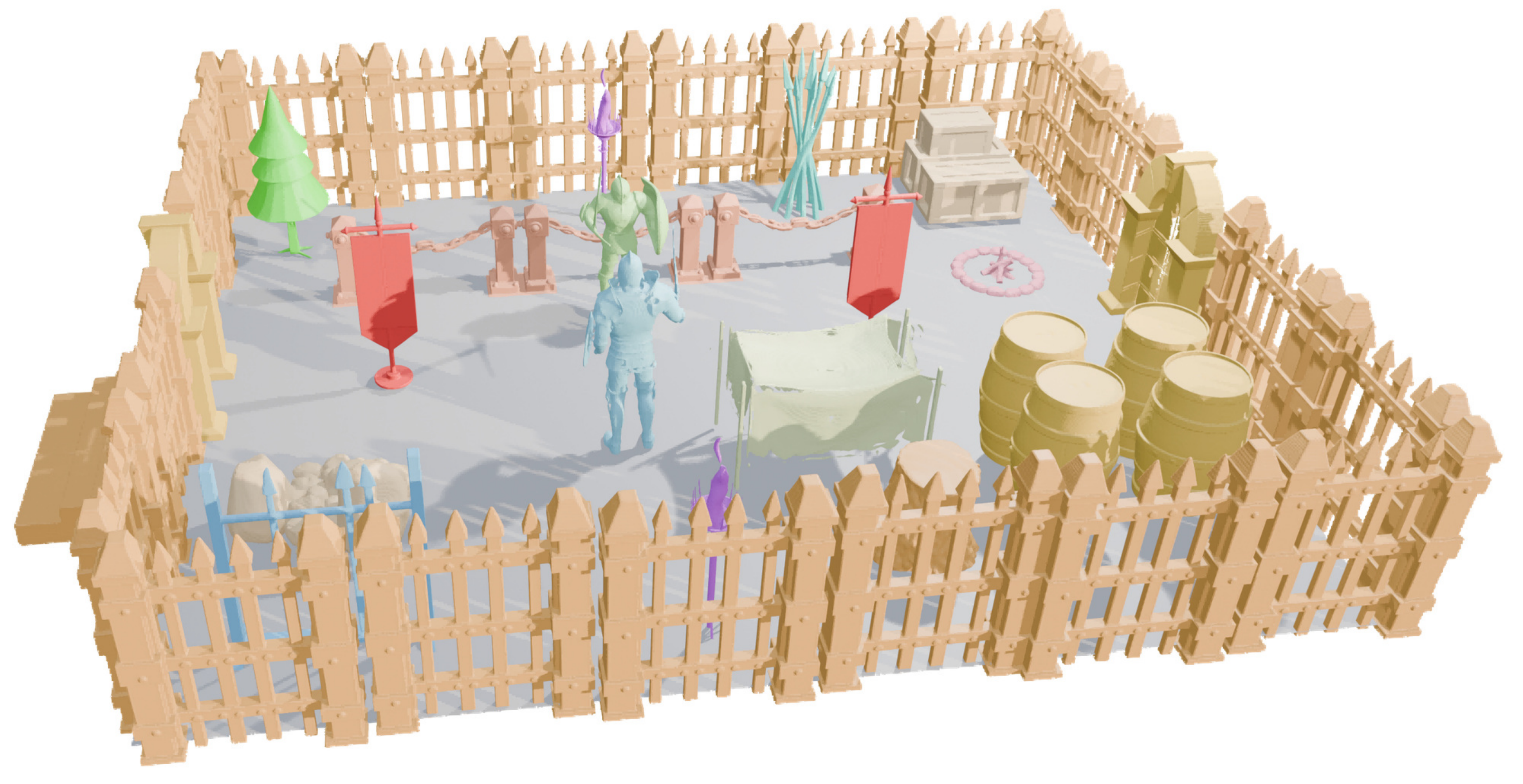}
    \vspace{0.16em}
    {\captionsetup{font=scriptsize,aboveskip=2pt,belowskip=0pt}
    \captionof{figure}{\textbf{Generated 3D assets.} Block3D meshes are manually assembled into an arena and rendered in Blender~\cite{blender}; prompts and isolated assets are documented in Appendix Section~D.1.}
    \label{fig:teaser}}
  \end{minipage}
  \end{center}
}
\ziplabendtitle{While text-to-3D generation has advanced rapidly, achieving high geometric fidelity at low inference cost remains challenging. Existing text-to-3D methods either decode discrete shape tokens autoregressively or iteratively refine global 3D representations with diffusion or flow models. However, autoregressive decoding is sequential and cannot revise errors, whereas diffusion and flow-matching models repeatedly process the full representation, making high-quality generation increasingly expensive. In this paper, we propose Block3D, a block-wise diffusion framework that partitions the discrete shape-token sequence into contiguous blocks, generates the blocks autoregressively, and jointly denoises all tokens within the current block. To alleviate error accumulation, we introduce confidence-guided intra-block correction, which revises low-confidence tokens before each block is finalized. On a held-out set from TRELLIS-500K, Block3D reduces mean end-to-end generation time from 25.71 seconds to 4.99 seconds, achieving a $5.15\times$ speedup over the fine-tuned autoregressive baseline without sacrificing geometric fidelity.
}

\clearpage
\section{Introduction}
 
High-quality 3D assets are increasingly required in game development, film production, virtual and augmented reality, robotics, and embodied AI, creating a growing demand for systems that can generate 3D content from natural-language descriptions. Recent methods have substantially improved the fidelity and diversity of generated assets by learning effective 3D representations~\cite{xiang2025trellis,zhao2025hunyuan3d2,yang2025pandora3d,bhat2025cube,ye2025shapellmomni}. Despite this progress, generating high-fidelity geometry typically requires autoregressive methods to sequentially decode long shape-token sequences or diffusion methods to repeatedly refine the complete latent representation over multiple denoising steps. As the representation becomes more detailed, both choices increase inference cost. Consequently, existing methods still struggle to achieve high geometric quality and low generation latency at the same time. Practical deployment further demands interactive latency, as highlighted by recent real-time 3D scene generation systems~\cite{ni2025wonderturbo}.

\begin{wrapfigure}{r}{0.49\textwidth}
    \vspace{-0.75\baselineskip}
    \centering
    \includegraphics[width=\linewidth]{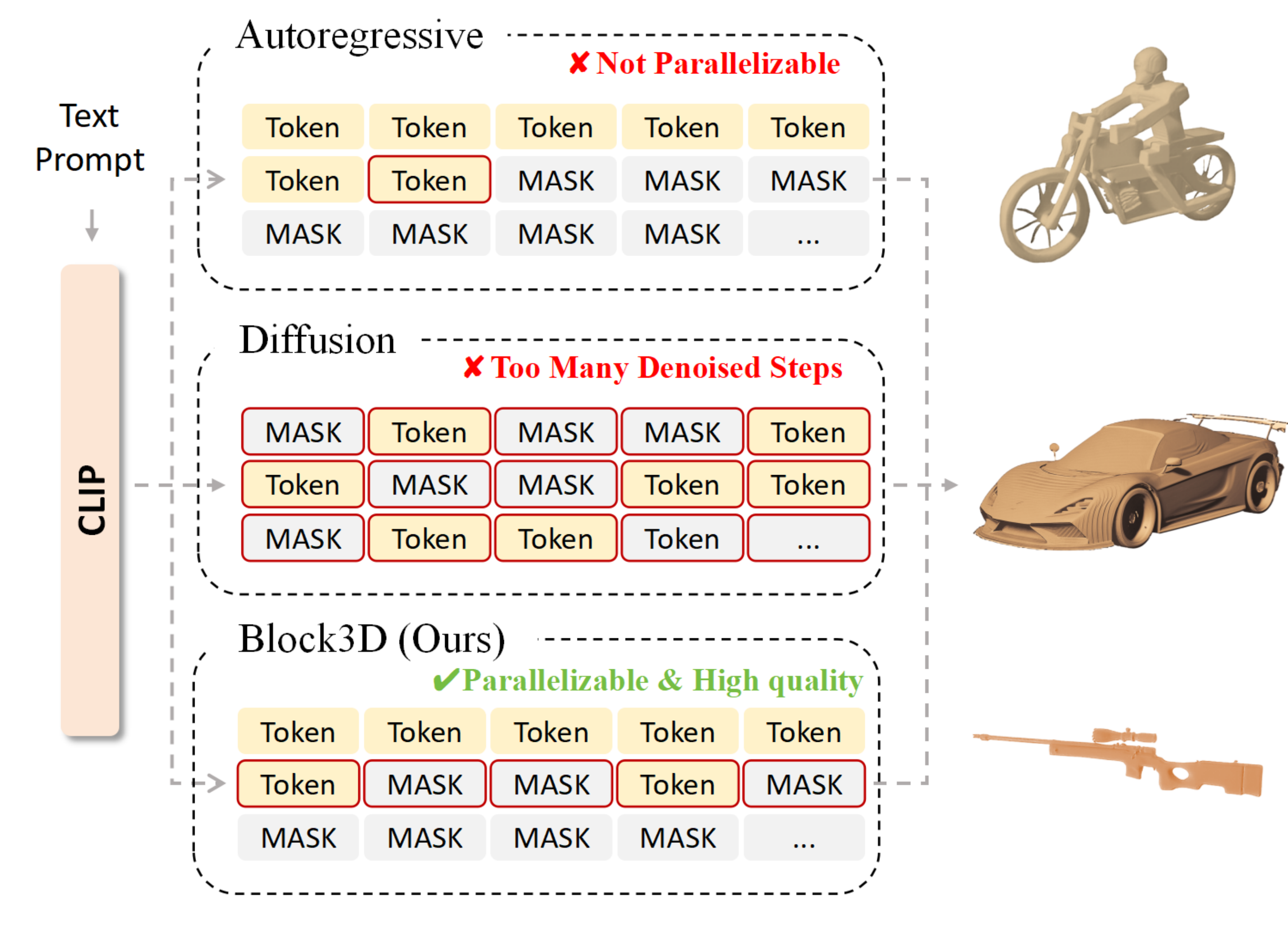}
    \captionsetup{width=\linewidth,font=footnotesize}
    \caption{\textbf{Comparison of 3D token generation schedules.}
    (a) Token-wise generation predicts one shape token at a time and cannot revise emitted tokens.
    (b) Full-sequence denoising updates all positions in parallel but repeatedly processes the complete representation.
    (c) Block3D generates blocks from left to right and denoises all positions in the current block in parallel; a filled token can still be corrected before its block is finalized.}
    \label{fig:concept}
    \vspace{-0.5\baselineskip}
\end{wrapfigure}

In this paper, we propose Block3D, a block-wise autoregressive diffusion framework for efficient text-to-3D generation. Its central idea is to shift the causal dependency of autoregressive generation from individual shape tokens to contiguous latent blocks. Blocks are generated causally from left to right, while all tokens within the current block are jointly denoised under bidirectional attention. To preserve geometric fidelity when predicting all tokens within a block in parallel, our confidence-guided correction mechanism not only fills masked positions but also revises low-confidence tokens before the current block is finalized. Consequently, Block3D substantially reduces end-to-end generation latency while alleviating error accumulation within each block. Experiments on 100 held-out objects show that Block3D achieves the best geometric metrics among the evaluated methods while reducing the mean end-to-end generation time from $25.71$ seconds to $4.99$ seconds compared with the separately fine-tuned Cube~\cite{bhat2025cube} baseline, achieving a $5.15\times$ speedup.

Recent work has pursued different efficiency strategies within diffusion/flow and autoregressive generation. Diffusion and flow models avoid token-wise decoding by refining multiple latent variables in parallel. TRELLIS, Hunyuan3D~2.0, and Pandora3D reduce the state being refined through compressed 3D latents, while TSSR denoises the complete mesh-token sequence and PartDiffuser limits each denoising stage to a semantic part~\cite{xiang2025trellis,zhao2025hunyuan3d2,yang2025pandora3d,song2025tssr,yang2026partdiffuser}. These designs reduce the representation size or the active denoising region, but detailed geometry still requires repeated refinement; full-sequence denoising revisits every token, whereas part-wise denoising additionally relies on point-cloud conditioning and semantic part segmentation. Autoregressive generators instead reduce sequential cost by organizing geometry into adaptive, hierarchical, or local units. Adaptive octree tokenization allocates tokens according to geometric complexity, PointNSP and OctGPT generate geometry progressively across spatial scales, TreeMeshGPT follows triangle adjacency, and MeshMosaic generates and assembles local mesh patches~\cite{deng2025octreetokenization,meng2025pointnsp,wei2025octgpt,lionar2025treemeshgpt,xu2025meshmosaic}. Although these representations shorten the causal sequence or its individual dependencies, fine geometry still requires additional tokens, hierarchy levels, branches, or patches to be generated sequentially. Consequently, high-fidelity generation continues to involve repeated latent refinement in diffusion-based methods or substantial sequential decoding in autoregressive methods, making low-latency generation difficult to achieve.

In summary, our main contributions are as follows:
\begin{itemize}
    \item We introduce Block3D, which shifts the causal dependency in Cube~\cite{bhat2025cube} from individual shape tokens to latent blocks and performs parallel denoising within each block, substantially reducing sequential generation latency.
    \item We introduce confidence-guided intra-block correction, which combines mask-to-token recovery with token-to-token editing~\cite{bie2026llada21} to alleviate error accumulation before the current block is finalized.
    \item Experiments on 100 held-out objects show that Block3D achieves the best geometric metrics among the evaluated methods and reduces mean end-to-end generation time from $25.71$ seconds to $4.99$ seconds relative to the fine-tuned Cube~\cite{bhat2025cube} baseline.
\end{itemize}

\section{Related Work}

\subsection{Diffusion-Based 3D Generation}
Recent feed-forward 3DGS work further studies scalable scene representations through compact latent compression~\cite{wang2025zpressor}; these image-conditioned systems are complementary to our text-conditioned mesh generator. Diffusion and flow matching support several distinct text-to-3D paradigms. DreamFusion, Magic3D, and MOC-3D optimize a 3D representation with supervision derived from pretrained 2D diffusion models~\cite{poole2022dreamfusion,lin2023magic3d,fan2026moc3d}. Shap-E and Rodin instead diffuse learned 3D representations, while TRELLIS performs flow matching over structured 3D latents~\cite{jun2023shap,wang2023rodin,xiang2025trellis}. Hunyuan3D~2.0, Pandora3D, and CraftsMan3D use staged or native 3D generation pipelines for detailed assets~\cite{zhao2025hunyuan3d2,yang2025pandora3d,wang2024craftsman}. LGM directly predicts 3D Gaussians from multi-view images, whereas DreamCS and VLM3D study learned or vision-language reward signals~\cite{tang2024lgm,zou2025dreamcs,bai2025vlm3d}. These systems should therefore not be grouped under a single VAE-plus-latent-diffusion formulation.

Recent work reduces optimization cost or improves view consistency through iterative reconstruction, flow distillation, direct trajectory design, and multi-view memory~\cite{chen2024microdreamer,yan2025consistent,li2025walking,zhou2025consdreamer,yang2025bridging,zhou2026stream3d}. These approaches update a global, implicit, or multi-view representation. Block3D instead retains a fixed discrete 3D representation and shortens the sequential horizon of its learned prior without changing the tokenizer or mesh decoder.

\subsection{AR-Based 3D Generation}
Another line of work formulates 3D generation as sequence prediction. MeshGPT autoregressively generates quantized triangle sequences, whereas Cube represents a shape by a compact, fixed-length sequence of VQ codes~\cite{siddiqui2024meshgpt,bhat2025cube}. MeshAnything, MeshAnything~V2, MeshRipple, and HiFi-Mesh improve mesh tokenization or shorten local autoregressive dependence~\cite{chen2024meshanything,chen2024meshanythingv2,lin2025meshripple,li2026hifimesh}. LLaMA-Mesh serializes vertices and faces as text tokens for a language model~\cite{wang2024llamamesh}. ShapeLLM targets language-grounded 3D understanding, while ShapeLLM-Omni extends multimodal language modeling to both understanding and generation~\cite{qi2024shapellm,ye2025shapellmomni}. VAR-3D introduces a view-aware 3D tokenizer, and AR3D-R1 studies reinforcement learning for autoregressive text-to-3D generation~\cite{han2025var3d,tang2025ar3dr1}.

Token-wise autoregression incurs one sequential decision per code and makes every emitted code irreversible. Later predictions are consequently conditioned on generated rather than ground-truth prefixes, the standard exposure-bias setting~\cite{bengio2015scheduled,ranzato2015sequence}. Block3D does not remove generated-prefix exposure bias: completed blocks remain fixed. It provides only a bounded opportunity to revise codes within the active block before that block becomes part of the prefix.

\subsection{Block-Wise and Editable Decoding}
Discrete denoising models replace continuous Gaussian noise with categorical transitions, while masked generators iteratively reveal token subsets with bidirectional context~\cite{austin2021structured,chang2022maskgit}. In 3D generation, PartDiffuser uses semantic mesh parts as causal groups and denoises tokens within each part, whereas TSSR performs global discrete mesh-token generation followed by remasking-based refinement~\cite{yang2026partdiffuser,song2025tssr}. Block3D differs from both by operating on part-free, fixed-length Cube~\cite{bhat2025cube} codes and freezing every completed prefix block.

Block Diffusion establishes autoregressive factorization across blocks, bidirectional denoising within the active block, clean/noisy training attention, and prefix caching~\cite{arriola2025blockdiffusion}. Subsequent systems study efficient block-diffusion language decoding, vision-language tokens, sparse attention, and vision-language-action generation~\cite{wu2025fastdllmv2,cheng2025sdarvl,xi2026losa,wang2026blockvla,lee2026tbdvla}. LLaDA2.1 introduces M2T/T2T editing and mixed mask/random-token supervision with multi-turn-forward augmentation~\cite{bie2026llada21}. Block3D adopts these established mechanisms but specifies the sample-level mixture, shape-code corruption, residual objective, confidence-gated updates, and deterministic quota for fixed-length 3D codes. The resulting T2T operation is restricted to the active block and executed within the fixed decoding horizon; Appendix Section~A.4 gives the complete update procedure.

\section{Method}
\label{sec:method}

\begin{figure}[t]
    \centering
    \includegraphics[width=\columnwidth]{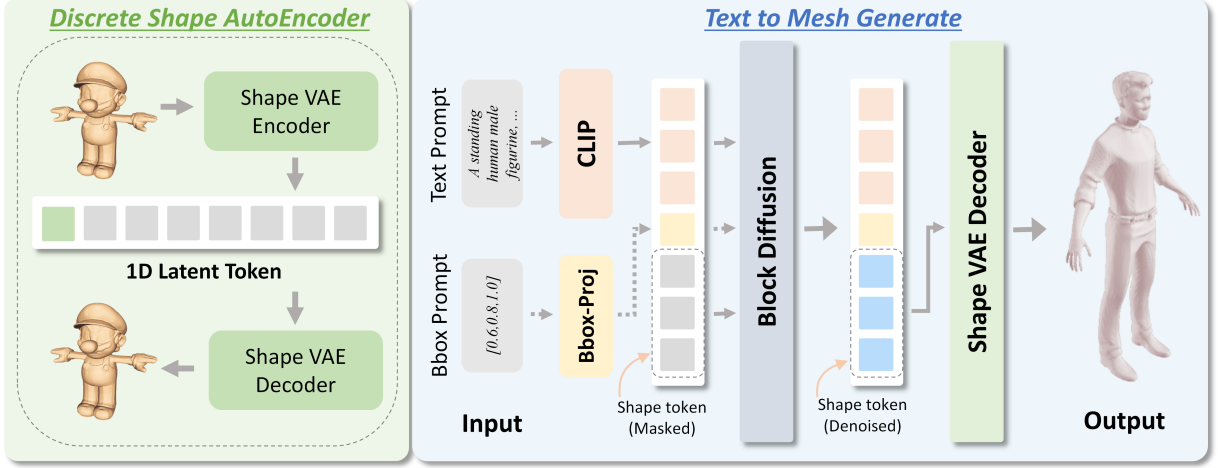}
    \caption{\textbf{Block3D overview.} Text and optional bounding-box conditions guide left-to-right block generation. Completed blocks form a frozen causal prefix, while M2T and T2T~\cite{bie2026llada21} edit only the active block before the frozen decoder maps the completed codes to a mesh.}
    \label{fig:pipeline}
\end{figure}

\subsection{Overview}
\label{sec:method_overview}

Figure~\ref{fig:pipeline} shows the complete text-to-mesh pipeline. We use the frozen VQ autoencoder of Cube~\cite{bhat2025cube}, which represents each mesh with $N=1024$ discrete codes from a codebook of size $V=16384$ and decodes a completed code sequence back to a mesh. Given a prompt $y$, a frozen CLIP ViT-L/14 text encoder~\cite{radford2021learning} produces 77 token-level features; an optional projected bounding-box token can be appended, although every reported experiment uses text alone. The resulting condition $C$ drives the Cube-initialized generator. The $N$ shape-code positions are divided into $K$ contiguous blocks. Generation follows the block-causal schedule of Block Diffusion~\cite{arriola2025blockdiffusion}: the current block starts fully masked, attends to the condition and the committed prefix, and predicts all active positions in parallel. Within that block, the M2T and T2T updates adapted from LLaDA2.1~\cite{bie2026llada21} fill masked positions and revise already filled positions before commitment. The completed block is then frozen and added to the cached prefix, and the same process continues from left to right. After all $K$ blocks are committed, the frozen VQ decoder converts the completed code sequence $\hat{x}$ into the output mesh $\hat{S}$.

The method has three components. \textbf{(1) Conditioned Block-Causal Denoising} partitions the fixed-length shape sequence into contiguous computational blocks, defines the clean/corrupted training visibility, and restricts bidirectional attention to the active block. \textbf{(2) Edit-Aware Training} exposes the generator to both masked and substituted shape codes, applies one model-based rollout, and supervises only the residual errors after that rollout. \textbf{(3) Bounded Confidence-Guided Decoding} uses conditional confidence for M2T filling and T2T replacement~\cite{bie2026llada21}, while a deterministic reveal quota guarantees that every mask is removed within at most $T$ iterations, where $T$ is the per-block update horizon. The first component transfers block-causal generation to fixed-length 3D codes; the latter two specify how editable states are trained and completed under a fixed inference horizon. All revisions remain intra-block because committed prefix blocks are never reopened.

\subsection{Block-Causal Shape-Code Denoising}
\label{sec:block_generation}

The conditional prior retains the DualStream RoFormer generator of Cube~\cite{bhat2025cube}. We write its shape-code logits as $\ell=F_\theta(C,z;A,p)$, where $z$ is the visible shape state, $A$ is its attention mask, and $p$ gives the logical position indices. Probabilities and generated codes use only the first $V$ output entries. The auxiliary mask $[M]$ reuses the inherited padding-token identifier for input embedding lookup, but it is excluded from output normalization and is never passed to the frozen shape decoder.

Let the block size $B$ and denoising horizon $T$ be positive integers. We form $K=\lceil N/B\rceil$ contiguous blocks. Block $k$ contains the positions $I_k=\{(k-1)B+1,\ldots,\min(kB,N)\}$ for $k=1,\ldots,K$, so only the final block can contain fewer than $B$ codes.
Writing $x^{(k)}=x_{I_k}$, $\hat{x}^{(k)}=\hat{x}_{I_k}$, and $\hat{x}^{(<k)}=(\hat{x}^{(1)},\ldots,\hat{x}^{(k-1)})$, the block-causal dependency inherited from Block Diffusion~\cite{arriola2025blockdiffusion} is realized as
\begin{equation}
\hat{x}^{(k)}=
\mathcal{G}_{\theta,T}\!\left(\hat{x}^{(<k)},C\right),
\qquad k=1,\ldots,K,
\label{eq:block_transition}
\end{equation}
where $\mathcal{G}_{\theta,T}$ is the deterministic transition in Algorithm~\ref{alg:block3d_sampling}; the other fixed decoding hyperparameters are suppressed from its notation. Equation~(\ref{eq:denoising_loss}) trains the token denoiser used inside this transition; we do not interpret $\mathcal{G}_{\theta,T}$ as an exact block likelihood.

At inference for block $k$, the generator receives $[\hat{x}^{(<k)};z_k]$ on the shape side. Prefix queries use token-causal visibility. Every active-block query attends to the complete prefix and bidirectionally to all positions in $z_k$; future blocks are absent.

Adapting the vectorized clean-and-corrupted construction of Block Diffusion~\cite{arriola2025blockdiffusion} to Cube~\cite{bhat2025cube}, training evaluates all block contexts in one forward pass over the concatenated sequence $[x;\tilde{x}]$. A query in the clean copy can attend only to clean tokens in its own block and all preceding blocks; it cannot attend to a later clean block or any token in the corrupted copy. A query in the corrupted copy can attend to the clean tokens of all preceding blocks and bidirectionally to the corrupted tokens within its own block. It cannot access the clean target tokens of the current block, clean tokens from future blocks, or corrupted tokens from any other block. Every shape query additionally attends to all condition tokens, whereas condition queries attend only to condition tokens. This visibility rule allows the corrupted states of all blocks to be trained in parallel without exposing the clean target of the block being reconstructed.

The two copies share logical positions, $p(x_i)=p(\tilde{x}_i)=i$. At inference, prefix queries use token-causal visibility and active-block queries remain bidirectional. Training thus uses block-bidirectional teacher-forced prefixes, whereas inference uses token-causal generated prefixes. Caching reduces repeated computation~\cite{arriola2025blockdiffusion}, but does not remove this mismatch or exposure bias. Appendix Section~A.2 gives the complete masks and position-ID construction.

\subsection{Edit-Aware Corruption and Training}
\label{sec:training}

LLaDA2.1~\cite{bie2026llada21} motivates supervision on both masked and substituted states but does not determine the shape-code corruption used here. For each training sample and block, we draw $\tau_k\sim\mathcal{U}(t_{\min},t_{\max})$ and independently sample $e_i\sim\mathrm{Bernoulli}(\tau_k)$ for $i\in I_k$. A single sample-level variable $a\sim\mathrm{Bernoulli}(\rho)$ selects the initial corruption stream: $a=0$ is M2T and $a=1$ is T2T. We set
\begin{equation}
\tilde{x}^{(0)}_i=
\left\{
\begin{array}{ll}
[M], & e_i=1,\ a=0,\\
u_i, & e_i=1,\ a=1,\\
x_i, & e_i=0,
\end{array}
\right.
\label{eq:training_corruption}
\end{equation}
where $u_i$ is uniform over the $V-1$ shape codes different from $x_i$. We use $(t_{\min},t_{\max})=(0.45,0.95)$ and $\rho=0.5$. The variables $\tau_k$ control corruption only and are not supplied to $F_\theta$ as time embeddings. Uniform wrong codes diversify substituted-state supervision, while the model-based rollout below introduces model-generated candidates before the residual loss. Inference starts each active block from the all-mask state.

Classifier-free condition dropout~\cite{ho2022classifierfree} is sampled once per batch element before either model evaluation. Let $\bar{C}_n=C_n$ for a retained condition and $\bar{C}_n=C_n^-$ for a dropped condition, where $C_n^-$ contains the empty-text encoding and, when enabled, a zero bounding box. The same realized condition $\bar{C}_n$ is reused by both evaluations below.

Inspired by the multi-turn-forward augmentation of LLaDA2.1~\cite{bie2026llada21}, we instantiate one model-based rollout ($R=1$) before computing the loss. For every block, the rollout initializes $z_k^{(0)}=\tilde{x}^{(0)}_{I_k}$. A no-gradient forward pass on $[x;\tilde{x}^{(0)}]$ supplies unguided logits ($g=0$) under $\bar{C}_n$, after which the update sets in Equations~(\ref{eq:m2t_set})--(\ref{eq:t2t_set}) are applied independently within every corrupted block using $(\eta_M,\eta_T)$ and the first-step quota $q_0$. Under the default $B=64$ and $T=4$, $q_0=16$. The simultaneous updates define $z_k^{(1)}$, and the rollout output is assembled by setting $\tilde{x}^{(R)}_{I_k}=z_k^{(1)}$ for every block. This state may contain masks, correct codes, and model-produced incorrect codes even when its initial state used only one corruption stream.

We then perform a separate gradient-carrying forward pass on $[x;\tilde{x}^{(R)}]$ and retain only corrupted-copy logits. For batch index $n$, let $\pi^{(R)}_{\theta,n,i}$ be their softmax over the first $V$ entries. Under the training visibility rule above, this distribution is conditioned on $\bar{C}_n$, the clean blocks preceding the block that contains position $i$, and the current corrupted block. Over all valid shape positions in the batch, define $\mathcal{D}=\{(n,i):\tilde{x}^{(R)}_{n,i}\neq x_{n,i}\}$. The objective is
\begin{equation}
\mathcal{L}=-\frac{1}{|\mathcal{D}|+\epsilon}
\sum_{(n,i)\in\mathcal{D}}
\log\pi^{(R)}_{\theta,n,i}(x_{n,i}).
\label{eq:denoising_loss}
\end{equation}
We set $\epsilon=10^{-8}$ and reduce over all residual tokens in the batch. Samples without residuals add no term; an empty $\mathcal{D}$ yields zero loss. The one-step rollout adds a generation-dependent state, while preceding blocks remain teacher-forced and cross-block exposure bias remains. Appendix Section~A.3 specifies corruption, rollout ordering, residual reduction, and empty-set handling.

\subsection{Bounded Confidence-Guided Decoding}
\label{sec:inference}

\begin{algorithm}[t]
\caption{Bounded editable block decoding}
\label{alg:block3d_sampling}
\small
\begin{algorithmic}[1]
\REQUIRE Conditional branch $C^+$; unconditional $C^-$ if $g>0$; $B,T,g,\eta_M,\eta_T$
\STATE Initialize $\hat{x}\leftarrow[M]^N$
\FOR{$k=1$ to $K$}
  \STATE Set $z_{k,i}^{(0)}\leftarrow[M]$ for every $i\in I_k$; set $S_k\leftarrow T$
  \STATE Build the batched condition/prefix cache
  \FOR{$s=0$ to $T-1$}
    \STATE Compute $\ell_s^+$ and, if $g>0$, $\ell_s^-$ from active state $z_k^{(s)}$
    \STATE Compute $\hat{z}_s,\alpha_s$ by Equations~(\ref{eq:guided_logits})--(\ref{eq:conditional_confidence})
    \STATE Compute $q_s$ and $\mathcal{M}_s$ from the current active state
    \STATE Form $\mathcal{U}^{\mathrm{M2T}}_s,\mathcal{U}^{\mathrm{T2T}}_s$ by Equations~(\ref{eq:m2t_set})--(\ref{eq:t2t_set})
    \IF{$\mathcal{M}_s=\emptyset$ and both update sets are empty}
      \STATE Set $S_k\leftarrow s$; \textbf{break}
    \ENDIF
    \STATE Compute $z_{k,i}^{(s+1)}$ by Equation~(\ref{eq:state_update}) for every $i\in I_k$
  \ENDFOR
  \STATE Commit $\hat{x}_{I_k}\leftarrow z_k^{(S_k)}$ and freeze block $k$
\ENDFOR
\RETURN the frozen shape decoder output $D_{\mathrm{shape}}(\hat{x})$
\end{algorithmic}
\end{algorithm}

\begin{figure*}[t]
    \centering
    \includegraphics[width=\textwidth]{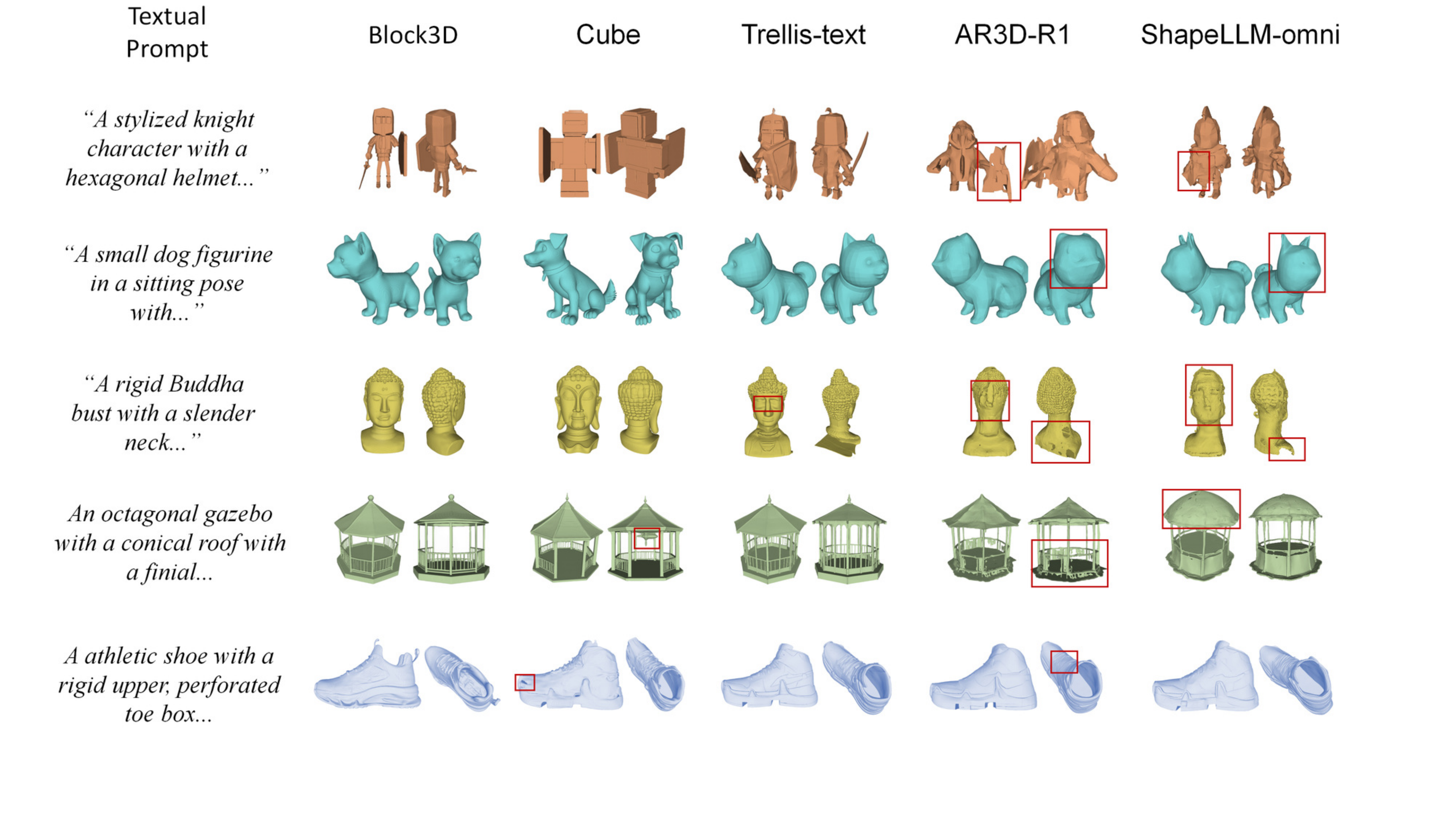}
    \caption{\textbf{Qualitative Comparison of Text-to-3D Generation.} Compared to existing paradigms, Block3D produces coherent front and back views in these examples, while several baselines exhibit missing or distorted geometry. Complete prompts and the selection protocol are documented in Appendix Section~D.2.}
    \label{fig:qualitative_comp}
\end{figure*}

Let the guidance coefficient satisfy $g\geq0$, and let $\eta_M,\eta_T\in[0,1]$ be the M2T and T2T confidence thresholds~\cite{bie2026llada21}. For block $k$, let $n_k=|I_k|$ and initialize $z_{k,i}^{(0)}=[M]$ for every $i\in I_k$. We suppress $k$ in iteration-level logits and update sets. At iteration $s\in\{0,\ldots,T-1\}$, the conditional branch $C^+$ encodes the prompt, while $C^-$ encodes the empty prompt and, when enabled, a zero bounding box. The branches produce $\ell_s^+$ and $\ell_s^-$. Classifier-free guidance (CFG)~\cite{ho2022classifierfree} gives
\begin{equation}
\ell_s^{g}=(1+\gamma_s)\ell_s^{+}-\gamma_s\ell_s^{-},
\qquad \gamma_s=g\frac{T-s}{T}.
\label{eq:guided_logits}
\end{equation}
Thus the guidance coefficient decreases from $g$ to $g/T$, rather than to zero; when $g=0$, the unconditional branch is omitted and $\ell_s^g=\ell_s^+$. For the deterministic sampler used in our experiments, the candidate and its acceptance confidence are
\begin{equation}
\begin{array}{rcl}
\hat{z}_{s,i}
&=&\mathop{\arg\max}_{0\leq v<V}\ell^g_{s,i,v},\\
\alpha_{s,i}
&=&\mathop{\mathrm{softmax}}(\ell^+_{s,i,0:V-1})[\hat{z}_{s,i}].
\end{array}
\label{eq:conditional_confidence}
\end{equation}
The guided logits propose a shape code, whereas the conditional logits determine whether it is accepted; CFG therefore does not directly inflate the threshold score. Ties in the argmax are resolved by selecting the smallest shape-code index.

We assign each iteration a deterministic minimum reveal quota
\begin{equation}
q_s=\left\lfloor\frac{n_k}{T}\right\rfloor
+\mathbf{1}\!\left[s<n_k\bmod T\right],
\qquad \sum_{s=0}^{T-1}q_s=n_k.
\label{eq:transfer_budget}
\end{equation}
Let $\mathcal{M}_s=\{i\in I_k:z_{k,i}^{(s)}=[M]\}$. Following the editable update in LLaDA2.1~\cite{bie2026llada21}, M2T and T2T select
\begin{equation}
\begin{array}{rcl}
\mathcal{U}^{\mathrm{M2T}}_s
&=&\{i\in\mathcal{M}_s:\alpha_{s,i}>\eta_M\}\\
&&{}\cup\ \mathop{\mathrm{TopK}}\limits_{i\in\mathcal{M}_s}
(\alpha_{s,i},\min(q_s,|\mathcal{M}_s|)),
\end{array}
\label{eq:m2t_set}
\end{equation}
\begin{equation}
\mathcal{U}^{\mathrm{T2T}}_s=
\{i\in I_k\setminus\mathcal{M}_s:\hat{z}_{s,i}\neq z_{k,i}^{(s)},\
\alpha_{s,i}>\eta_T\}.
\label{eq:t2t_set}
\end{equation}
Here, $\mathrm{TopK}_{i\in\mathcal{M}_s}(\alpha_{s,i},r)$ returns the $r$ masked positions with the largest confidence values and returns the empty set when $r=0$. Confidence ties are resolved in ascending order of the global position index $i$. This defines short final blocks and settings with $n_k<T$. Appendix Section~A.4 gives the complete sampler, including tie handling and short-block execution.
The active state is updated simultaneously:
\begin{equation}
z_{k,i}^{(s+1)}=
\left\{
\begin{array}{ll}
\hat{z}_{s,i}, & i\in\mathcal{U}^{\mathrm{M2T}}_s\cup
\mathcal{U}^{\mathrm{T2T}}_s,\\
z_{k,i}^{(s)}, & \mbox{otherwise.}
\end{array}
\right.
\label{eq:state_update}
\end{equation}
High-confidence M2T updates can make the number of revealed positions exceed $q_s$, so $q_s$ is a lower bound on progress rather than an upper update budget. T2T does not inspect the confidence of the old code: it replaces a filled code only when a different new candidate exceeds $\eta_T$.

If $m_s=|\mathcal{M}_s|$, the TopK fallback reveals at least $\min(q_s,m_s)$ masked positions, so Equation~(\ref{eq:m2t_set}) implies $m_{s+1}\leq\max(m_s-q_s,0)$. Applying this bound across iterations and using $\sum_{s=0}^{T-1}q_s=n_k$ gives $m_T=0$. The quota therefore completes every active block within $T$ iterations while confidence-gated T2T editing~\cite{bie2026llada21} remains available at every iteration before commitment.

We retain the 23-layer DualStream RoFormer from Cube~\cite{bhat2025cube} (12 heads, width 1536), bypass its inherited single-stream layer on this path, and initialize the first $V$ token embeddings from a projection of the frozen VQ codebook. We fine-tune only this generator. Inference uses $B=64$, $T=4$, $(\eta_M,\eta_T)=(0.95,0.9)$ for M2T/T2T editing~\cite{bie2026llada21}, and guidance coefficient $g=3.0$ under Equation~(\ref{eq:guided_logits}), giving $K=16$.

With CFG~\cite{ho2022classifierfree}, conditional and unconditional branches are concatenated in the batch. Decoding makes $K$ cache calls and at most $KT$ active-block calls, equivalent to at most $2K(T+1)$ logical branch evaluations when $g>0$; the default uses at most 80 batched or 160 logical calls. Following Block Diffusion~\cite{arriola2025blockdiffusion}, the condition and prefix cache are reused within the $T$ updates and rebuilt after commitment. Thus intra-block revision has a fixed horizon, while completed blocks cannot be corrected. Appendix Section~A.5 gives the cache construction and model-call complexity derivation.

\section{Experiments}

\subsection{Experimental Settings}

\paragraph{Datasets.}
We fine-tune the text-to-shape generator on 300K objects sampled from TRELLIS-500K~\cite{xiang2025trellis}, a large-scale collection of 3D assets with paired text descriptions. To construct the evaluation set, we first randomly select 100 objects from TRELLIS-500K using seed 42 and exclude these exact objects from the fine-tuning candidate pool. Each shape is converted into discrete shape tokens using the frozen Cube~\cite{bhat2025cube} tokenizer, and its paired text serves as the generation condition. Each evaluation object retains its paired text prompt, and every method generates one shape per prompt. Appendix Section~B.1 records the split, identifiers, prompts, and exclusion rule; Appendix Section~B.2 specifies preprocessing.

\paragraph{Baselines and Evaluation Metrics.}

\begin{table}[t]
\centering
\caption{Geometry and text-shape alignment on the 100-object TRELLIS-500K~\cite{xiang2025trellis} evaluation set. F@1 denotes an F-score threshold of 1\%~\cite{fan2017pointset,mescheder2019occupancy}. Geometry metrics are rounded to three decimals and CLIPScore~\cite{hessel2021clipscore} to two decimals.}
\label{tab:main_results}
\begingroup
\small
\renewcommand{\arraystretch}{1.10}
\begin{tabular}{@{}lcccc@{}}
\toprule
Method & CD-L1$\downarrow$ & NC$\uparrow$ & F@1\%$\uparrow$ & CLIP$\uparrow$ \\
\midrule
ShapeLLM-Omni & 0.229 & 0.490 & 0.089 & 19.08 \\
TRELLIS-text  & 0.222 & 0.496 & 0.090 & 20.41 \\
AR3D-R1       & 0.145 & 0.583 & 0.162 & 21.92 \\
Cube          & 0.094 & 0.632 & 0.219 & \textbf{23.87} \\
Block3D       & \textbf{0.078} & \textbf{0.668} & \textbf{0.309} & 23.24 \\
\bottomrule
\end{tabular}
\endgroup
\end{table}

\begin{table}[t]
\centering
\caption{End-to-end generation time in seconds over the 100 TRELLIS-500K~\cite{xiang2025trellis} prompts, measured on one NVIDIA A100 80GB GPU.}
\label{tab:efficiency}
\begingroup
\small
\renewcommand{\arraystretch}{1.10}
\begin{tabular}{@{}lcccc@{}}
\toprule
Method & Mean$\downarrow$ & Median$\downarrow$ & P90$\downarrow$ & Std.$\downarrow$ \\
\midrule
TRELLIS-text  & 11.65 & 9.75  & 20.06 & 5.59 \\
ShapeLLM-Omni & 36.89 & 33.69 & 40.64 & 8.48 \\
AR3D-R1       & 79.80 & 82.48 & 94.19 & 17.34 \\
Cube          & 25.71 & 25.43 & 26.68 & 0.80 \\
Block3D       & \textbf{4.99} & \textbf{4.96} & \textbf{5.60} & \textbf{0.43} \\
\bottomrule
\end{tabular}
\endgroup
\end{table}

We compare Block3D with ShapeLLM-Omni~\cite{ye2025shapellmomni}, TRELLIS-text~\cite{xiang2025trellis}, AR3D-R1~\cite{tang2025ar3dr1}, and Cube~\cite{bhat2025cube}. Cube provides the controlled comparison because both models use the same released checkpoint, frozen geometry components, training subset, and optimization budget; the remaining methods use their official inference settings. Appendix Section~C.4 lists releases and complete baseline settings.

Geometry evaluation samples 8,192 surface points and normals per mesh pair and reports Chamfer-$L_1$ distance, normal consistency, and F-score at 1\% of the target bounding-box diagonal~\cite{fan2017pointset,mescheder2019occupancy}. Eight-view CLIPScore measures text-shape alignment with CLIP ViT-L/14~\cite{hessel2021clipscore,radford2021learning}. Latency statistics are measured on one NVIDIA A100 and include condition encoding, shape-code generation, and mesh decoding, but exclude model loading and disk I/O. Protocol details appear in Appendix Section~C.1 (geometry), C.2 (CLIPScore), C.3 (latency), and C.5 (invalid outputs and statistical scope).

\paragraph{Implementation Details.}
We initialize Block3D from the released Cube~\cite{bhat2025cube} checkpoint and freeze its shape tokenizer, CLIP ViT-L/14~\cite{radford2021learning} text encoder, and shape decoder. Only the text-to-shape generator is fine-tuned. We train for 35K steps with AdamW~\cite{loshchilov2019decoupled}, a learning rate of $1\times10^{-4}$, bfloat16 precision, and a global batch size of 40 on four NVIDIA A100 80GB GPUs. We use a 100-step linear warm-up, zero weight decay, a gradient clipping norm of 1.0, and random seed 42. Appendix Section~B.3 gives the complete training and checkpoint configuration.

During training, the block corruption level is sampled uniformly from $[0.45,0.95]$, the T2T branch~\cite{bie2026llada21} is selected with probability $0.5$, and one no-gradient model-based filling rollout is performed. Classifier-free condition dropout~\cite{ho2022classifierfree} is applied with probability $0.1$. Unless otherwise stated, inference uses a block size of 64, four denoising steps per block, M2T and T2T confidence thresholds of 0.95 and 0.9, and guidance coefficient $g=3.0$ in Equation~(\ref{eq:guided_logits}). With $T=4$, $\gamma_s$ decreases linearly from 3.0 to 0.75 across the four iterations. We use deterministic argmax decoding without top-$p$ sampling. Parameter selection is in Appendix Section~B.4; code and materials are indexed in Appendix Sections~E.1--E.3.

\subsection{Comparison with 3D Generation Methods}

\paragraph{Geometry Metrics Comparison.}
Table~\ref{tab:main_results} compares paired geometric fidelity and text-shape alignment. Under the shared evaluation protocol, Block3D attains the strongest CD-L1, NC, and F@1\% scores~\cite{fan2017pointset,mescheder2019occupancy} among the evaluated methods, while its CLIPScore~\cite{hessel2021clipscore} remains competitive. The controlled comparison with Cube~\cite{bhat2025cube} shows that block-wise denoising improves paired geometric fidelity in addition to accelerating generation.

\begin{table}[t]
\centering
\caption{Effect of block size on Block3D with four denoising steps per block. Time is reported in seconds. Every configuration is trained independently.}
\label{tab:block_ablation}
\begingroup
\small
\renewcommand{\arraystretch}{1.10}
\begin{tabular}{@{}lcccc@{}}
\toprule
Model & Time$\downarrow$ & CD-L1$\downarrow$ & NC$\uparrow$ & F@1\%$\uparrow$ \\
\midrule
Block3D ($B=32$)  & 12.98 & \textbf{0.076} & 0.667 & 0.296 \\
Block3D ($B=64$)  & 4.99  & 0.078 & \textbf{0.668} & \textbf{0.309} \\
Block3D ($B=96$)  & 3.62  & 0.085 & 0.660 & 0.279 \\
Block3D ($B=128$) & 2.90  & 0.090 & 0.647 & 0.258 \\
Block3D ($B=256$) & \textbf{2.15} & 0.186 & 0.578 & 0.103 \\
\bottomrule
\end{tabular}
\endgroup
\end{table}

\begin{table}[t]
\centering
\caption{Effect of denoising steps on Block3D with a block size of 64. Time is reported in seconds. Every configuration is trained independently.}
\label{tab:step_ablation}
\begingroup
\small
\renewcommand{\arraystretch}{1.10}
\begin{tabular}{@{}lcccc@{}}
\toprule
Model & Time$\downarrow$ & CD-L1$\downarrow$ & NC$\uparrow$ & F@1\%$\uparrow$ \\
\midrule
Block3D ($T=4$)  & \textbf{4.99} & \textbf{0.078} & 0.668 & \textbf{0.309} \\
Block3D ($T=8$)  & 7.33  & 0.079 & 0.665 & 0.300 \\
Block3D ($T=12$) & 9.37  & 0.084 & 0.668 & 0.284 \\
Block3D ($T=20$) & 13.66 & \textbf{0.078} & \textbf{0.676} & 0.303 \\
\bottomrule
\end{tabular}
\endgroup
\end{table}

\begin{table}[t]
\centering
\caption{Effect of token-to-token editing~\cite{bie2026llada21}. Geometry metrics are rounded to three decimals and CLIPScore~\cite{hessel2021clipscore} to two decimals.}
\label{tab:t2t_ablation}
\begingroup
\small
\renewcommand{\arraystretch}{1.10}
\begin{tabular}{@{}lcccc@{}}
\toprule
Model & CD-L1$\downarrow$ & NC$\uparrow$ & F@1\%$\uparrow$ & CLIP$\uparrow$ \\
\midrule
Block3D (M2T)       & 0.081 & 0.666 & 0.287 & 22.74 \\
Block3D (M2T+T2T)   & \textbf{0.078} & \textbf{0.668} & \textbf{0.309} & \textbf{23.24} \\
\bottomrule
\end{tabular}
\endgroup
\end{table}

\paragraph{Efficiency Metrics Comparison.}
Table~\ref{tab:efficiency} shows that Block3D is the fastest evaluated method across all reported latency statistics. Its mean end-to-end generation time is 4.99 seconds, corresponding to a $5.15\times$ speedup over the controlled Cube~\cite{bhat2025cube} baseline. The consistent improvements in median and P90 latency show that this advantage holds across the 100 TRELLIS-500K~\cite{xiang2025trellis} prompts rather than arising from a small number of easy cases.

\subsection{Ablation Study}

We study block size, the number of denoising steps, and token-to-token editing~\cite{bie2026llada21}. Every configuration in Tables~\ref{tab:block_ablation} and~\ref{tab:step_ablation}, as well as both variants in Table~\ref{tab:t2t_ablation}, is trained independently for the same number of optimization steps and evaluated on the same 100-object TRELLIS-500K~\cite{xiang2025trellis} evaluation set.

\paragraph{Effect of Block Size.}
Table~\ref{tab:block_ablation} shows that larger blocks reduce the number of left-to-right stages but make parallel denoising harder. Moving from $B=64$ to 96 lowers latency from 4.99 to 3.62 seconds but reduces F@1\% from 0.309 to 0.279; at $B=256$, F@1\% falls to 0.103. We therefore use $B=64$ as the quality--latency balance.

\paragraph{Effect of Denoising Steps.}
Table~\ref{tab:step_ablation} shows that additional steps increase latency without monotonic quality gains. At $T=20$, latency rises to 13.66 seconds while CD-L1 remains 0.078 and F@1\% is 0.303, compared with 4.99 seconds and 0.309 at $T=4$. We therefore use four steps.

\paragraph{Effect of Token-to-Token Editing.}
Table~\ref{tab:t2t_ablation} compares the complete model with an independently trained M2T-only variant~\cite{bie2026llada21}. Adding T2T editing reduces CD-L1 by 4.7\%, raises F@1\% from 0.287 to 0.309, and raises CLIPScore from 22.74 to 23.24. The simultaneous gains support token correction for revising errors that mask-only recovery cannot address.

\subsection{Discussion and Limitations}

Block3D combines the frozen Cube representation~\cite{bhat2025cube}, the Block Diffusion schedule~\cite{arriola2025blockdiffusion}, and LLaDA2.1 editing~\cite{bie2026llada21} with shape-code corruption, residual rollout supervision, and confidence-guided active-block revision. Because blocks follow Cube's one-dimensional code order and committed prefixes remain immutable, the method targets intra-block errors rather than semantic-part structure or cross-block exposure bias. Its fixed horizon bounds decoding cost and completes every active block within $T$ steps (Appendix Section~A.4).

\section{Conclusion}

Block3D replaces the token-wise prior of Cube~\cite{bhat2025cube} with block-causal denoising, generating blocks from left to right while jointly denoising and correcting the active shape codes. Among the evaluated methods, it improves paired geometry and reduces Cube's mean generation time from 25.71 to 4.99 seconds while retaining competitive text-shape alignment. Completed blocks remain fixed; future work will study cross-block refinement and broader shape representations and datasets.

\clearpage
\bibliographystyle{unsrtnat}
\bibliography{reference}

\clearpage
\section*{Appendix}

\appendix
\makeatletter
\@addtoreset{equation}{section}
\@addtoreset{algorithm}{section}
\makeatother
\setcounter{equation}{0}
\setcounter{figure}{0}
\setcounter{table}{0}
\setcounter{algorithm}{0}
\renewcommand{\theequation}{\thesection.\arabic{equation}}
\renewcommand{\thefigure}{\Alph{figure}}
\renewcommand{\thetable}{\Alph{table}}
\renewcommand{\thealgorithm}{\thesection.\arabic{algorithm}}
\providecommand{\masktoken}{[M]}



\section{Complete Method Details}
\label{sec:complete_method}

\subsection{Frozen Shape Representation and Generator}
\label{sec:frozen_shape}

\paragraph{Shape Representation.}
Block3D retains the frozen vector-quantized shape representation of Cube~\cite{bhat2025cube}. A mesh $S$ is encoded as a fixed-length sequence
\begin{equation}
\begin{aligned}
x&=(x_1,\ldots,x_N), \qquad N=1024,\\
x_i&\in\{0,\ldots,V-1\}.
\end{aligned}
\label{eq:shape_codes}
\end{equation}
with codebook size $V=16384$. The Cube shape encoder is used to prepare shape-code supervision before generator training. The corresponding shape decoder $D_{\mathrm{shape}}$ remains frozen and converts only a completed sequence of valid shape codes into the output mesh. The auxiliary mask symbol $\masktoken$ reuses the inherited padding-token identifier for input embedding lookup, but it is excluded from the $V$-way output normalization and is never passed to the shape decoder.

\paragraph{Condition Representation.}
Given a text prompt, the frozen CLIP ViT-L/14 encoder~\cite{radford2021learning} produces 77 token-level condition features. Cube also supports an optional projected bounding-box token. All experiments reported in the main paper use text alone, so no bounding-box token is appended in the reported training or inference runs. Classifier-free condition dropout replaces the retained text condition with the frozen empty-text encoding; if bounding-box conditioning is enabled in another setting, its unconditional counterpart is a zero bounding box.

\paragraph{Trainable Generator.}
The conditional shape prior is initialized from Cube's 23-layer DualStream RoFormer generator~\cite{bhat2025cube}, with 12 attention heads and hidden width 1536. Block3D uses the dual-stream path and bypasses the inherited single-stream layer. The first $V$ input token embeddings are initialized from a learned projection of the frozen VQ codebook. During fine-tuning, the VQ shape encoder, VQ codebook, CLIP text encoder, and shape decoder remain frozen; only the text-to-shape generator is updated. At inference, all components are frozen.

\subsection{Training Attention and Position IDs}
\label{sec:attention_positions}

\paragraph{Block Notation.}
For block size $B$, the $N$ shape positions are divided into $K=\lceil N/B\rceil$ contiguous blocks. Block $r$ contains
\begin{equation}
I_r=\{(r-1)B+1,\ldots,\min(rB,N)\}.
\label{eq:block_indices}
\end{equation}
Training concatenates a clean sequence $x$ and a corrupted sequence $\tilde{x}$, while keeping a single condition sequence $C$. We write $x^{(r)}=x_{I_r}$ and $\tilde{x}^{(r)}=\tilde{x}_{I_r}$.

\paragraph{Complete Visibility Rule.}
The attention mask adapts the vectorized clean/noisy construction of Block Diffusion~\cite{arriola2025blockdiffusion} to fixed-length Cube codes. Let $A(q,k)=1$ indicate that query $q$ can attend to key $k$. Condition queries are isolated from the shape sequence:
\begin{equation}
A(C,C)=1,\qquad A(C,x^{(s)})=A(C,\tilde{x}^{(s)})=0.
\label{eq:condition_attention}
\end{equation}
Every shape query can attend to all condition tokens. A clean query in block $r$ attends block-bidirectionally to its own clean block and to every preceding clean block, but never to the corrupted copy:
\begin{equation}
\begin{aligned}
A(x^{(r)},C)&=1,\\
A(x^{(r)},x^{(s)})&=\mathbf{1}[s\leq r],\\
A(x^{(r)},\tilde{x}^{(s)})&=0.
\end{aligned}
\label{eq:clean_attention}
\end{equation}
A corrupted query in block $r$ attends to clean blocks strictly before $r$ and bidirectionally to the corrupted tokens in its own block:
\begin{equation}
\begin{aligned}
A(\tilde{x}^{(r)},C)&=1,\\
A(\tilde{x}^{(r)},x^{(s)})&=\mathbf{1}[s<r],\\
A(\tilde{x}^{(r)},\tilde{x}^{(s)})&=\mathbf{1}[s=r].
\end{aligned}
\label{eq:corrupt_attention}
\end{equation}
Consequently, a corrupted query cannot access the clean target tokens of its own block, any future clean block, or a corrupted state from another block. All corrupted block contexts can nevertheless be evaluated in one forward pass because each block receives its own clean teacher-forced prefix.

\paragraph{Logical Position IDs.}
The clean and corrupted copies share logical shape positions:
\begin{equation}
p(x_i)=p(\tilde{x}_i)=i, \qquad i=1,\ldots,N.
\label{eq:logical_positions}
\end{equation}
Thus duplication for training changes the physical sequence layout but not the RoFormer position assigned to a shape code. At inference, a committed prefix position $\hat{x}_i$ retains its global ID $i$, and an active-block state $z_{k,i}$ also uses global ID $i$. Prefix queries use token-causal visibility, whereas active-block queries attend to the complete prefix and bidirectionally within the active block. Future blocks are absent from the inference sequence.

\begin{figure}[t]
\centering
\includegraphics[width=0.96\textwidth]{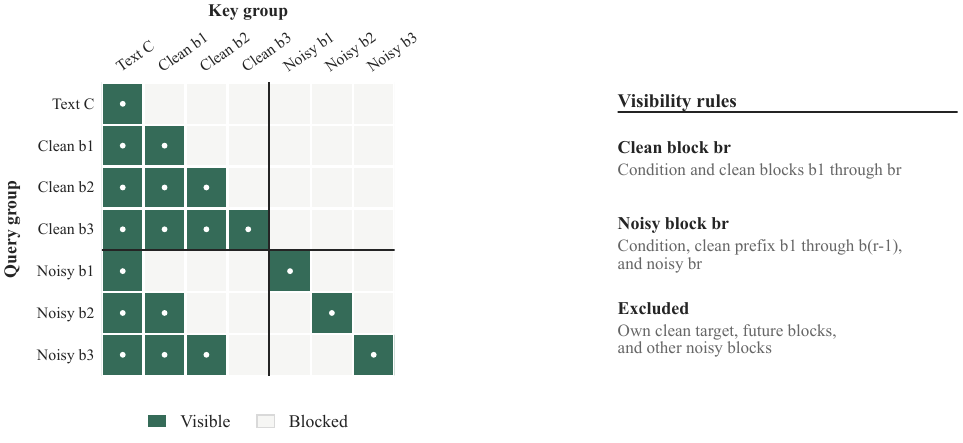}
\caption{Block-level visualization of the training attention mask. Green cells denote visible query--key group pairs. A corrupted query reads the condition, its clean teacher-forced prefix, and the corrupted copy of its own block; it cannot read the clean target of that block, a future block, or another corrupted block. The diagram expands each block to a single group for readability; the corresponding token-level rules are given in Equations~\ref{eq:condition_attention}--\ref{eq:corrupt_attention}.}
\label{fig:attention_visibility}
\end{figure}

\paragraph{Training--Inference Distinction.}
The training construction supplies block-bidirectional, teacher-forced clean prefixes, while inference conditions on token-causal, model-generated prefixes. Prefix caching reduces repeated computation but does not remove this exposure difference. Completed blocks are immutable in both the cache and the generated sequence.

Figure~\ref{fig:attention_visibility} makes the leakage boundary explicit. Bidirectional visibility is confined to the matching block, while the clean prefix remains strictly earlier than the corrupted target block. The duplicated clean and corrupted streams therefore vectorize supervision over all block indices without exposing any block to its own target codes.

\subsection{Corruption, Rollout, and Residual Loss}
\label{sec:corruption_rollout}

\paragraph{Sample-Level Corruption Stream.}
For each training sample, a single variable $a\sim\mathrm{Bernoulli}(\rho)$ chooses either mask-to-token (M2T, $a=0$) or token-to-token (T2T, $a=1$) corruption for the complete shape sequence. For every block $k$, a corruption rate $\tau_k\sim\mathcal{U}(t_{\min},t_{\max})$ is drawn independently, followed by $e_i\sim\mathrm{Bernoulli}(\tau_k)$ for $i\in I_k$. The initial state is
\begin{equation}
\tilde{x}^{(0)}_i=
\begin{cases}
\masktoken, & e_i=1,\ a=0,\\
u_i, & e_i=1,\ a=1,\\
x_i, & e_i=0,
\end{cases}
\label{eq:supp_corruption}
\end{equation}
where $u_i$ is sampled uniformly from the $V-1$ codes different from $x_i$. The reported setting uses $(t_{\min},t_{\max})=(0.45,0.95)$ and $\rho=0.5$. The sampled corruption rate changes only the input state and is not provided to the generator as a time embedding.

\paragraph{Condition Dropout.}
Classifier-free condition dropout~\cite{ho2022classifierfree} is sampled once per batch element with probability 0.1. The realized retained or dropped condition is reused for both the rollout forward pass and the gradient-carrying loss forward pass, preventing the two passes from receiving inconsistent conditions.

\paragraph{One-Step Model Rollout.}
Block3D uses one model-based rollout ($R=1$), inspired by the multi-turn-forward augmentation of LLaDA2.1~\cite{bie2026llada21}. Starting from $z_k^{(0)}=\tilde{x}^{(0)}_{I_k}$, an unguided, no-gradient forward pass evaluates all corrupted blocks with the attention mask in Section~\ref{sec:attention_positions}. Each block then applies the M2T and T2T update rules from Section~\ref{sec:complete_sampler} using the first reveal quota $q_0$. Under $B=64$ and $T=4$, $q_0=16$. All selected positions are updated simultaneously to obtain $z_k^{(1)}$, and these per-block states are reassembled into $\tilde{x}^{(R)}$. The rollout output may contain masks, correct codes, and model-produced incorrect codes regardless of which initial corruption stream was selected.

\paragraph{Residual Objective.}
A separate gradient-carrying forward pass evaluates $[x;\tilde{x}^{(R)}]$. Let $\pi^{(R)}_{\theta,n,i}$ be the softmax over the first $V$ logits for batch item $n$ and position $i$, and define the residual set
\begin{equation}
\mathcal{D}=\{(n,i):\tilde{x}^{(R)}_{n,i}\neq x_{n,i}\}.
\label{eq:residual_set}
\end{equation}
The training loss is
\begin{equation}
\mathcal{L}=-\frac{1}{|\mathcal{D}|+\epsilon}
\sum_{(n,i)\in\mathcal{D}}
\log \pi^{(R)}_{\theta,n,i}(x_{n,i}),
\qquad \epsilon=10^{-8}.
\label{eq:residual_loss}
\end{equation}
Reduction is performed over all residual tokens in the batch rather than averaging separate per-sample losses. A sample with no residual positions contributes no term to the numerator. If the entire batch has $\mathcal{D}=\emptyset$, the numerator is empty and the batch loss is zero. Padding or otherwise invalid shape positions are excluded before $\mathcal{D}$ is formed.

\begin{algorithm}[H]
\caption{Edit-Aware Training State Construction}
\label{alg:training_state}
\begin{algorithmic}[1]
\REQUIRE Clean codes $x$; condition $C$; $B,T,t_{\min},t_{\max},\rho,\eta_M,\eta_T$
\STATE Sample condition dropout once for each batch item
\STATE Sample one stream variable $a$ for each batch item
\FOR{each block $k$}
  \STATE Sample $\tau_k$ and construct $\tilde{x}^{(0)}_{I_k}$ by Equation~\ref{eq:supp_corruption}
\ENDFOR
\STATE Run one no-gradient, unguided forward pass on $[x;\tilde{x}^{(0)}]$
\STATE Apply one simultaneous M2T/T2T update independently in every block
\STATE Assemble the updated blocks into $\tilde{x}^{(R)}$
\STATE Run a separate gradient-carrying forward pass on $[x;\tilde{x}^{(R)}]$
\STATE Form $\mathcal{D}$ and compute Equation~\ref{eq:residual_loss}
\RETURN Residual denoising loss $\mathcal{L}$
\end{algorithmic}
\end{algorithm}

\subsection{Complete Inference Algorithm}
\label{sec:complete_sampler}

\paragraph{Guided Proposal and Conditional Confidence.}
For active block $k$, let $z_{k,i}^{(0)}=\masktoken$ for $i\in I_k$ and $n_k=|I_k|$. At iteration $s\in\{0,\ldots,T-1\}$, the conditional and unconditional branches produce $\ell_s^+$ and $\ell_s^-$. Classifier-free guidance gives
\begin{equation}
\ell_s^g=(1+\gamma_s)\ell_s^+-\gamma_s\ell_s^-,
\qquad \gamma_s=g\frac{T-s}{T}.
\label{eq:supp_cfg}
\end{equation}
When $g=0$, the unconditional branch is omitted. The guided logits propose a candidate, while the conditional logits score its acceptance:
\begin{equation}
\begin{aligned}
\hat{z}_{s,i}&=\mathop{\arg\max}_{0\leq v<V}\ell^g_{s,i,v},\\
\alpha_{s,i}&=\mathrm{softmax}(\ell^+_{s,i,0:V-1})[\hat{z}_{s,i}].
\end{aligned}
\label{eq:supp_candidate}
\end{equation}
An argmax tie is resolved by choosing the smallest shape-code index.

\paragraph{Reveal Quota and Update Sets.}
Iteration $s$ receives the deterministic minimum reveal quota
\begin{equation}
q_s=\left\lfloor\frac{n_k}{T}\right\rfloor
+\mathbf{1}[s<n_k\bmod T],
\qquad \sum_{s=0}^{T-1}q_s=n_k.
\label{eq:supp_quota}
\end{equation}
Let $\mathcal{M}_s=\{i\in I_k:z_{k,i}^{(s)}=\masktoken\}$. The two update sets adapted from LLaDA2.1~\cite{bie2026llada21} are
\begin{equation}
\begin{aligned}
\mathcal{U}^{\mathrm{M2T}}_s={}&
\{i\in\mathcal{M}_s:\alpha_{s,i}>\eta_M\}\\
&\cup\ \mathrm{TopK}_{i\in\mathcal{M}_s}
\bigl(\alpha_{s,i},\min(q_s,|\mathcal{M}_s|)\bigr),
\end{aligned}
\label{eq:supp_m2t}
\end{equation}
\begin{equation}
\mathcal{U}^{\mathrm{T2T}}_s=
\{i\in I_k\setminus\mathcal{M}_s:
\hat{z}_{s,i}\neq z_{k,i}^{(s)},\ \alpha_{s,i}>\eta_T\}.
\label{eq:supp_t2t}
\end{equation}
For TopK confidence ties, the smaller global shape position is selected first. TopK returns the empty set when its requested size is zero. The union in Equation~\ref{eq:supp_m2t} means that confidence-qualified M2T updates may exceed $q_s$; the quota is a lower bound on reveal progress, not an upper bound on the number of updates.

\paragraph{Simultaneous Update and Completion.}
All accepted positions are updated from the same iteration state:
\begin{equation}
z_{k,i}^{(s+1)}=
\begin{cases}
\hat{z}_{s,i}, & i\in\mathcal{U}^{\mathrm{M2T}}_s\cup\mathcal{U}^{\mathrm{T2T}}_s,\\
z_{k,i}^{(s)}, & \text{otherwise.}
\end{cases}
\label{eq:supp_update}
\end{equation}
If $m_s=|\mathcal{M}_s|$, the fallback term reveals at least $\min(q_s,m_s)$ positions. Therefore
\begin{equation}
m_{s+1}\leq\max(m_s-q_s,0).
\label{eq:mask_bound}
\end{equation}
Together with $m_0=n_k$ and $\sum_s q_s=n_k$, Equation~\ref{eq:mask_bound} gives $m_T=0$. Every block is consequently complete after at most $T$ iterations. For a short block with $n_k<T$, the first $n_k$ quotas equal one and the remaining quotas equal zero; the same argument applies without padding the block to size $B$.

\begin{algorithm}[H]
\caption{Complete bounded editable block sampler}
\label{alg:complete_sampler}
\begin{algorithmic}[1]
\REQUIRE Condition $C^+$; optional $C^-$; $N,V,B,T,g,\eta_M,\eta_T$
\STATE Set $K\leftarrow\lceil N/B\rceil$ and initialize $\hat{x}\leftarrow\masktoken^N$
\FOR{$k=1$ to $K$}
  \STATE Set $z_k^{(0)}\leftarrow\masktoken^{n_k}$ and $S_k\leftarrow T$
  \STATE Build the batched condition/prefix cache for block $k$
  \FOR{$s=0$ to $T-1$}
    \STATE Evaluate $\ell_s^+$ and, if $g>0$, $\ell_s^-$
    \STATE Compute $\hat{z}_s$ and $\alpha_s$ by Equations~\ref{eq:supp_cfg}--\ref{eq:supp_candidate}
    \STATE Form $\mathcal{M}_s$, $q_s$, $\mathcal{U}^{\mathrm{M2T}}_s$, and $\mathcal{U}^{\mathrm{T2T}}_s$
    \IF{$\mathcal{M}_s=\emptyset$ and both update sets are empty}
      \STATE Set $S_k\leftarrow s$; \textbf{break}
    \ENDIF
    \STATE Update all active positions simultaneously by Equation~\ref{eq:supp_update}
  \ENDFOR
  \STATE Commit $\hat{x}_{I_k}\leftarrow z_k^{(S_k)}$ and freeze block $k$
\ENDFOR
\RETURN $D_{\mathrm{shape}}(\hat{x})$
\end{algorithmic}
\end{algorithm}

The reported sampler uses deterministic argmax decoding, without top-$p$ sampling, with $B=64$, $T=4$, $(\eta_M,\eta_T)=(0.95,0.9)$, and $g=3.0$. Because $N=1024$, this setting has $K=16$ full blocks and no short final block. The per-iteration guidance coefficients are $(3.0,2.25,1.5,0.75)$.

\subsection{Prefix Cache and Complexity}
\label{sec:cache_complexity}

\paragraph{Cache Construction.}
At the start of block $k$, the condition and committed prefix $\hat{x}^{(<k)}$ are encoded for the conditional branch and, when $g>0$, for the unconditional branch. The two branches are concatenated along the batch dimension. Their condition and prefix key/value states remain unchanged across the $T$ active-block updates and are therefore reused. Only the active-block states are recomputed after an M2T/T2T update. Once block $k$ is committed, its codes become part of the immutable prefix and the cache is rebuilt for block $k+1$.

\paragraph{Model-Call Accounting.}
For $K$ blocks and at most $T$ updates per block, decoding uses $K$ cache-building calls and at most $KT$ active-block calls. Conditional and unconditional branches are executed in the same batched call, so the batched-call upper bound is
\begin{equation}
K+KT=K(T+1).
\label{eq:batched_calls}
\end{equation}
When $g>0$, each batched call contains two logical branch evaluations, giving at most
\begin{equation}
2K(T+1)
\label{eq:logical_calls}
\end{equation}
logical evaluations. With $N=1024$, $B=64$, and $T=4$, the upper bounds are 80 batched calls and 160 logical branch evaluations. Early stopping can reduce active-block calls but does not increase either bound. This accounting measures generator invocations; it does not include the final frozen shape-decoder call.

\begin{figure}[t]
\centering
\includegraphics[width=0.96\textwidth]{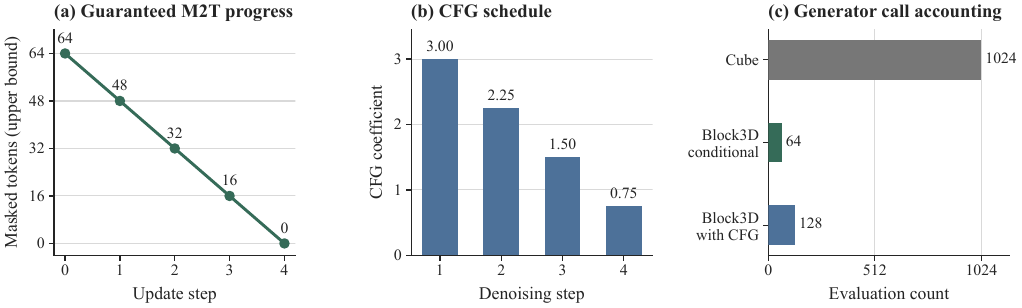}
\caption{Analytical diagnostics for the main configuration. Left: the deterministic quota reveals at least 16 positions per step, so the number of masks is bounded by $64\rightarrow48\rightarrow32\rightarrow16\rightarrow0$. Middle: Equation~\ref{eq:supp_cfg} yields guidance coefficients $(3.00,2.25,1.50,0.75)$. Right: excluding the $K$ cache-building calls, Block3D performs at most $KT=64$ conditional active-block evaluations, or $2KT=128$ logical branches with CFG, whereas token-wise Cube performs 1,024 sequential token evaluations. These counts describe the generator schedule and are not substitutes for measured end-to-end latency.}
\label{fig:schedule_accounting}
\end{figure}

\section{Data and Training Reproducibility}
\label{sec:data_training}

\subsection{Data Split and Prompt Provenance}
\label{sec:data_split}

The source pool is formed from TRELLIS-500K records~\cite{xiang2025trellis}, each of which associates a 3D asset with a paired text description. The paired text is used as the text-to-shape condition. Each evaluation object retains its paired prompt, and the same stored prompt is provided to every generation method without method-specific editing.

Using seed 42, we sample 100 valid asset records for evaluation and assign the stable indices 0000--0099. The recovered manifest contains 100 distinct prompts and 100 distinct source-mesh paths: 53 records from Objaverse-XL Sketchfab, 46 from Objaverse-XL GitHub, and one from ABO, all distributed through TRELLIS-500K. Every evaluated method receives this same ordered prompt list and produces one output for each prompt. Deterministic decoders use their fixed argmax path, while stochastic external baselines use seed 42.

Before fine-tuning, each evaluation record is matched against the candidate pool by its normalized pair-record path and, when that key is unavailable, by its normalized source-mesh path. Split construction stops if an evaluation record cannot be matched or if the number of excluded source records differs from 100. This enforces exact asset-level disjointness between the evaluation list and the fine-tuning pool; it does not claim category-level or semantic near-duplicate removal. The accompanying Code and Data Package includes \texttt{random100\_evaluation\_manifest.jsonl}, which records the stable index, fixed prompt, source collection, normalized TRELLIS-500K mesh path, pair-record key, and bounding-box vector for every evaluation item. The package identifies rather than redistributes the third-party target meshes; they are resolved from the referenced TRELLIS-500K snapshot.

To characterize the fixed list without introducing a hand-assigned category taxonomy, we report properties that can be recomputed directly from the ordered manifest. For target bounding-box vector $b_j=(b_{j,x},b_{j,y},b_{j,z})$, define the scale-invariant anisotropy
\begin{equation}
a_j=\frac{\max_d b_{j,d}}{\min_d b_{j,d}}.
\label{eq:bbox_anisotropy}
\end{equation}
Prompt length ranges from 10 to 38 whitespace-delimited words, with median 19.0. The median $a_j$ is 2.15, and 36 of 100 targets have $a_j>3$, so the paired set includes a substantial fraction of elongated or flattened shapes rather than only near-isotropic objects. These bounding-box vectors are used only by the common evaluator; no evaluated generator receives them as input.

\begin{figure}[t]
\centering
\includegraphics[width=0.96\textwidth]{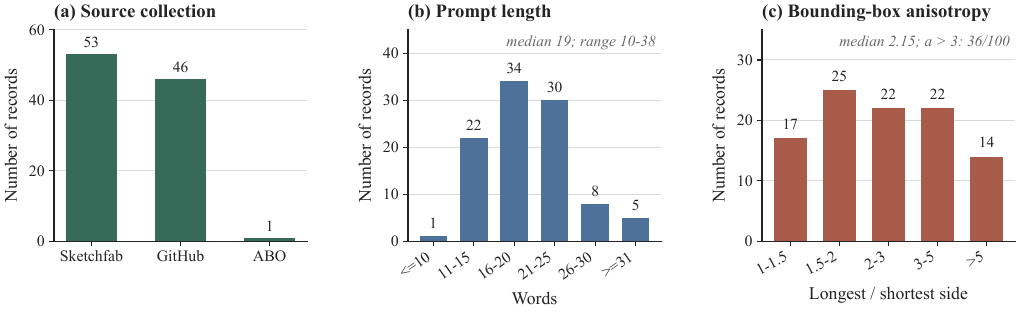}
\caption{Recomputable profile of the fixed random100 evaluation manifest. The source counts sum to 100; prompt-length bins use the stored text verbatim; and anisotropy uses Equation~\ref{eq:bbox_anisotropy}. The distribution reflects the seeded TRELLIS-500K sample and is not manually balanced by source, prompt length, or shape extent.}
\label{fig:evaluation_profile}
\end{figure}

\subsection{Data Preprocessing}
\label{sec:data_preprocessing}

We apply the released Cube preprocessing~\cite{bhat2025cube} before tokenization. Meshes are triangulated, non-finite geometry and zero-area faces are removed, and unreferenced vertices are discarded. All remaining nonempty connected components are retained. The original orientation is preserved: no category-specific rotation, canonical-pose fitting, or ICP alignment is applied. Each valid mesh is centered at its axis-aligned bounding-box midpoint and uniformly rescaled by its longest box side to fit the normalized Cube coordinate domain.

Every preprocessed training mesh and paired evaluation target is converted into a length-1024 discrete shape-code sequence with the same frozen Cube tokenizer. The sequence is cached together with its TRELLIS identifier and provides the clean target $x$ used by the corruption process in Section~\ref{sec:corruption_rollout}. An input is excluded only when preprocessing leaves no finite triangle surface. At inference, the frozen shape decoder receives a sequence only after all 1024 positions contain valid codebook entries. This single tokenizer keeps the representation fixed across Block3D training, the controlled Cube baseline, and paired targets; no learned geometry component is updated during Block3D fine-tuning.

\subsection{Training Configuration}
\label{sec:training_configuration}

We initialize from the released Cube checkpoint and optimize only the text-to-shape generator on 300K training objects. Training runs for 35K optimizer updates in bfloat16 on four NVIDIA A100 80GB GPUs with an effective global batch size of 40 and seed 42. AdamW~\cite{loshchilov2019decoupled} uses a learning rate of $10^{-4}$, $(\beta_1,\beta_2)=(0.9,0.95)$, $\epsilon=10^{-8}$, and zero weight decay. The learning rate increases linearly from zero to $10^{-4}$ during the first 100 updates and remains constant thereafter, and gradients are clipped to global norm 1.0 before each optimizer update. Classifier-free condition dropout is applied with probability 0.1.

The training corruption range is $[0.45,0.95]$, the T2T stream is selected with probability 0.5, and the main model uses one no-gradient model-based rollout. The VQ tokenizer, codebook, CLIP ViT-L/14 text encoder, and shape decoder remain frozen, and the 35K-step checkpoint is used for evaluation. All block-size, denoising-step, and M2T-only variants start from the same released Cube checkpoint and are independently trained for the same 35K-update budget.

The update and batch settings determine the amount of training data presented to the optimizer. Let $S=35{,}000$ be the number of updates, $G=4$ the number of workers, $b=10$ the local batch size, and $M=300{,}000$ the number of training records. Then
\begin{equation}
M_{\mathrm{seen}}=SGb=1.4\times10^6,
\qquad \frac{M_{\mathrm{seen}}}{M}\approx4.67.
\label{eq:training_presentations}
\end{equation}
Here $M_{\mathrm{seen}}$ counts sample presentations, not unique objects; shuffling is renewed by the distributed sampler. For the default full block, let $\mathcal{D}^{(0)}_k$ contain the initially corrupted positions and let $n_{\mathrm{T2T}}$ count T2T-stream samples in one global batch. The initial corruption process gives
\begin{equation}
\begin{aligned}
\mathbb{E}[\tau_k]&=\frac{0.45+0.95}{2}=0.70,\\
\mathbb{E}[|\mathcal{D}^{(0)}_k|]&=64(0.70)=44.8,\\
\mathbb{E}[n_{\mathrm{T2T}}]&=0.5(40)=20.
\end{aligned}
\label{eq:training_expectations}
\end{equation}
Thus an average initial block retains roughly 19.2 clean anchors before the model-based rollout, while an expected 20 samples in each global batch begin from the T2T corruption stream. These are analytical expectations under the sampled corruption process, not additional empirical measurements.

The reference software environment is Linux with Python 3.10 and PyTorch 2.2.2 or later built for CUDA. Training uses four NVIDIA A100 80GB GPUs, and the latency protocol in Section~\ref{sec:latency_protocol} uses one A100 80GB GPU on the same host for every method. The host CPU performs data loading, mesh import, and process orchestration, but CPU throughput is not a reported comparison variable. The released trainer records the operating-system platform, CPU model and logical core count, host memory, Python and PyTorch versions, CUDA runtime, GPU names, and GPU memory in \texttt{training\_run.json} for each reproduced run.

\paragraph{Optimization and Checkpointing.}
Gradient accumulation is one, so every loaded global batch produces one optimizer update. Under bfloat16 autocast, no float16 loss scaler is active. Gradients are checked for a finite global norm, clipped to 1.0, and then passed to AdamW; the optimizer is cleared with \texttt{set\_to\_none=True} before the next update. EMA is disabled. The main configuration performs no periodic validation, sample generation, or geometry evaluation during optimization, preventing an auxiliary evaluation loop from selecting the reported checkpoint. Model-only checkpoints are written every 5K updates, and the final 35K checkpoint is the one used by the reported evaluation.

\subsection{Parameter Selection}
\label{sec:parameter_selection}

The reported operating point is $B=64$, $T=4$, $(\eta_M,\eta_T)=(0.95,0.9)$, guidance coefficient $g=3.0$, corruption range $[0.45,0.95]$, T2T stream probability 0.5, and one rollout step. Block size controls the number of left-to-right stages, while $T$ controls the fixed number of active-block refinement opportunities.

The block-size study trains $B\in\{32,64,96,128,256\}$ with $T=4$. Increasing $B$ reduces latency but eventually weakens paired geometric fidelity; $B=64$ is used as the reported quality--latency balance. The denoising-step study fixes $B=64$ and trains $T\in\{4,8,12,20\}$. Additional steps increase measured latency without providing monotonic gains in the reported geometry metrics, so the main configuration uses $T=4$. The T2T study independently trains an M2T-only variant and the complete M2T+T2T model under the same optimization budget.

The remaining values are fixed globally rather than adjusted per prompt. The M2T/T2T thresholds $(0.95,0.9)$ follow the conservative public decoding defaults of LLaDA2.1~\cite{bie2026llada21}; the guidance coefficient $g=3.0$ is applied through the within-block decay in Equation~\ref{eq:supp_cfg}. The corruption interval $[0.45,0.95]$ exposes the model to substantially incomplete blocks while retaining clean anchors in most training states. One rollout is the lowest-cost nontrivial model-based augmentation and introduces model-generated errors without multiplying the gradient-carrying training pass. These settings are shared by the block-size and denoising-step ablations unless the corresponding variable is under study; the M2T-only comparison uses its dedicated M2T training configuration while retaining the same initialization and optimization budget.

\section{Complete Evaluation Protocol}
\label{sec:evaluation_protocol}

\subsection{Geometry Metrics}
\label{sec:geometry_metrics}

Generated and target meshes are loaded through the same geometry routine. Non-finite values, degenerate and duplicate faces, and unreferenced vertices are removed. Each mesh is centered at its own axis-aligned bounding-box midpoint and uniformly rescaled so that its longest box side equals 1.92, the Cube evaluation domain. This transform preserves aspect ratio and orientation; no rotational alignment, ICP, or prompt-specific registration is applied. From each transformed mesh, 8,192 points are sampled with probability proportional to triangle area; the evaluator uses base seed 0 and adds the stable record index, so record $j$ uses NumPy seed $j$ for its paired surface draws. Each point receives the unit face normal of its sampled triangle. Let $P=\{(p_i,n_i)\}$ and $Q=\{(q_j,m_j)\}$ denote the generated and target samples. The unsquared symmetric Chamfer distance is
\begin{equation}
\begin{aligned}
\mathrm{CD\mbox{-}L1}(P,Q)={}&
\frac{1}{2|P|}\sum_{p\in P}\min_{q\in Q}\|p-q\|_2\\
&+\frac{1}{2|Q|}\sum_{q\in Q}\min_{p\in P}\|q-p\|_2.
\end{aligned}
\label{eq:cd_l1}
\end{equation}
For each point, let $\nu_Q(p)$ be its nearest sampled target point and $\nu_P(q)$ the nearest generated point. Normal consistency is the symmetric mean of the absolute normal dot products:
\begin{equation}
\begin{aligned}
\mathrm{NC}(P,Q)={}&
\frac{1}{2|P|}\sum_{p\in P}|n_p^\top n_{\nu_Q(p)}|\\
&+\frac{1}{2|Q|}\sum_{q\in Q}|m_q^\top m_{\nu_P(q)}|.
\end{aligned}
\label{eq:normal_consistency}
\end{equation}
Let $b_Q$ be the target bounding-box vector stored with the paired TRELLIS-500K record after Cube normalization, and let $d_Q=\lVert b_Q\rVert_2$. All 100 reported records contain this vector; the evaluator falls back to the transformed target extent only when it is absent. With $\delta=0.01d_Q$, precision is the fraction of generated samples within $\delta$ of the target, recall is the fraction of target samples within $\delta$ of the generation, and
\begin{equation}
F@1\%=\frac{2\,\mathrm{precision}\,\mathrm{recall}}
{\mathrm{precision}+\mathrm{recall}}.
\label{eq:fscore}
\end{equation}
The main paper reports CD-L1, NC, and F@1\% to three decimals. Nearest neighbors are computed in Euclidean distance on the 8,192-point sets, and the same neighbors are used for the corresponding normal terms. These definitions follow standard point-sampled geometry evaluation~\cite{fan2017pointset,mescheder2019occupancy}.

The implementation samples positive-area triangles proportionally to area and draws uniform barycentric coordinates with the square-root transform. Record $j$ initializes one NumPy generator with seed $j$; target and generated surfaces are drawn sequentially from it. Two SciPy \texttt{cKDTree} queries compute both nearest-neighbor directions. The evaluator uses unsquared distances, reuses the same indices for normals, and saves per-sample values and status fields before macro-averaging.

\subsection{CLIPScore}
\label{sec:clipscore}

Text--shape alignment is evaluated from eight fixed views using CLIP ViT-L/14~\cite{radford2021learning}. Each mesh is centered and isotropically normalized by its axis-aligned bounding box, assigned the same neutral-gray material, and rendered at $512\times512$ pixels with PyTorch3D from eight azimuths separated by $45^\circ$. Camera elevation, distance, field of view, background, and lighting are fixed and shared by all methods. For prompt feature $t$ and image feature $v_r$ from view $r$, the score is
\begin{equation}
\mathrm{CLIPScore}(t,S)=\frac{100}{8}\sum_{r=1}^{8}
\max\!\left(\frac{t^\top v_r}{\|t\|_2\|v_r\|_2},0\right).
\label{eq:clipscore}
\end{equation}
The official CLIP preprocessing center-crops each view and applies the released image normalization. Prompts use the same 77-token truncation and text preprocessing as the frozen condition encoder. Equation~\ref{eq:clipscore} is the positive cosine form of CLIPScore~\cite{hessel2021clipscore}; scores are averaged over views and then over the 100 prompts and reported to two decimals.

\subsection{Latency}
\label{sec:latency_protocol}

Latency is measured in inference mode with batch size one on one NVIDIA A100 80GB GPU. Each model remains resident on the device, untimed warm-up runs are completed before measurement, and CUDA is synchronized immediately before and after every timed generation. The same ordered set of 100 TRELLIS-500K prompts is used for every method, with one timed generation per prompt. End-to-end time includes condition encoding, shape-code generation, and mesh decoding; model loading, checkpoint transfer, visualization, and disk I/O are excluded. Mean, median, P90, and standard deviation are computed over the 100 recorded generations.

Specifically, let $t_{(1)}\leq\cdots\leq t_{(100)}$ be the sorted per-prompt times and let $\bar{t}$ denote their mean. The recorded summaries use
\begin{equation}
\begin{aligned}
\bar{t}&=\frac{1}{100}\sum_{j=1}^{100}t_j,
&\mathrm{median}(t)&=\frac{t_{(50)}+t_{(51)}}{2},\\
\sigma_t&=\sqrt{\frac{1}{100}\sum_{j=1}^{100}(t_j-\bar{t})^2},
&\mathrm{P90}(t)&=t_{(90)}.
\end{aligned}
\label{eq:latency_statistics}
\end{equation}
The standard deviation is therefore the population standard deviation over the fixed evaluation list, and P90 uses the nearest-rank definition. Timing records remain paired with stable sample IDs so aggregate values can be traced back to individual prompts.

Block3D and Cube run in bfloat16 and share the same frozen Cube geometry components. Block3D uses $B=64$, $T=4$, $g=3.0$, and deterministic argmax decoding. The external methods retain their official inference precision, sampling schedule, kernels, and native mesh decoders. All mesh-decoding stages execute on the timed device and are included in the reported end-to-end latency. This measured protocol is separate from the analytical generator-call accounting in Section~\ref{sec:cache_complexity}.

\subsection{Baseline Settings}
\label{sec:baseline_settings}

\paragraph{Controlled Baseline.}
Cube~\cite{bhat2025cube} is the controlled baseline because it shares the released initialization, shape tokenizer, codebook, CLIP text encoder, shape decoder, 300K training subset, and 35K-step optimization budget with Block3D. The generator objective and decoding schedule differ. Block3D uses the block-causal editable decoder described in Section~\ref{sec:complete_sampler}, with $B=64$, $T=4$, and $g=3.0$.

\paragraph{External Baselines.}
ShapeLLM-Omni~\cite{ye2025shapellmomni}, TRELLIS-text~\cite{xiang2025trellis}, and AR3D-R1~\cite{tang2025ar3dr1} are external text-to-3D baselines. They use their official inference pipelines. Their native representations, sampling schedules, and postprocessing are retained rather than forced into the Cube pipeline.

For TRELLIS-text, we use the released \texttt{TRELLIS-text-xlarge} pipeline with 25 sparse-structure steps, 25 structured-latent steps, guidance scale 7.5 for both stages, seed 42, and its native mesh decoder. ShapeLLM-Omni and AR3D-R1 use their released text-to-3D checkpoints, default decoding parameters, and native mesh extraction. Cube uses greedy autoregressive decoding of all 1,024 codes. Block3D uses the deterministic sampler in Algorithm~\ref{alg:complete_sampler} for the quality evaluation. No method receives the target mesh, a bounding box, or method-specific prompt editing, and no method-specific geometry cleanup is applied before the common evaluation transform.

\subsection{Statistical Scope and Evaluation Boundaries}
\label{sec:statistical_scope}

The evaluation contains 100 paired prompts. Each method generates one shape per prompt under the same single-generation budget. Geometry and CLIP values are macro-averaged over the 100 ordered outputs, so each prompt contributes equal weight; latency is summarized over the same 100 prompts by its mean, median, P90, and standard deviation. Because every method is evaluated on the same prompt--target pairs, per-prompt comparisons are paired even though the main tables report aggregate values.

CD-L1, NC, and F@1\% quantify fidelity to the paired reference, while eight-view CLIPScore measures reference-free prompt alignment. An output is valid when it contains finite vertices, at least one nondegenerate face, and a nonzero bounding-box extent after native mesh extraction. Generation, geometry evaluation, and CLIP evaluation succeeded for all 100 records in every row reported in the main quality table; therefore no missing-value imputation or failure penalty affects the reported means. The evaluator records generation success, mesh validity, and metric status for each sample and does not regenerate a failed output.

\section{Qualitative Results and Prompts}
\label{sec:qualitative}
\begin{figure}[p]
\centering
\includegraphics[width=0.82\textwidth]{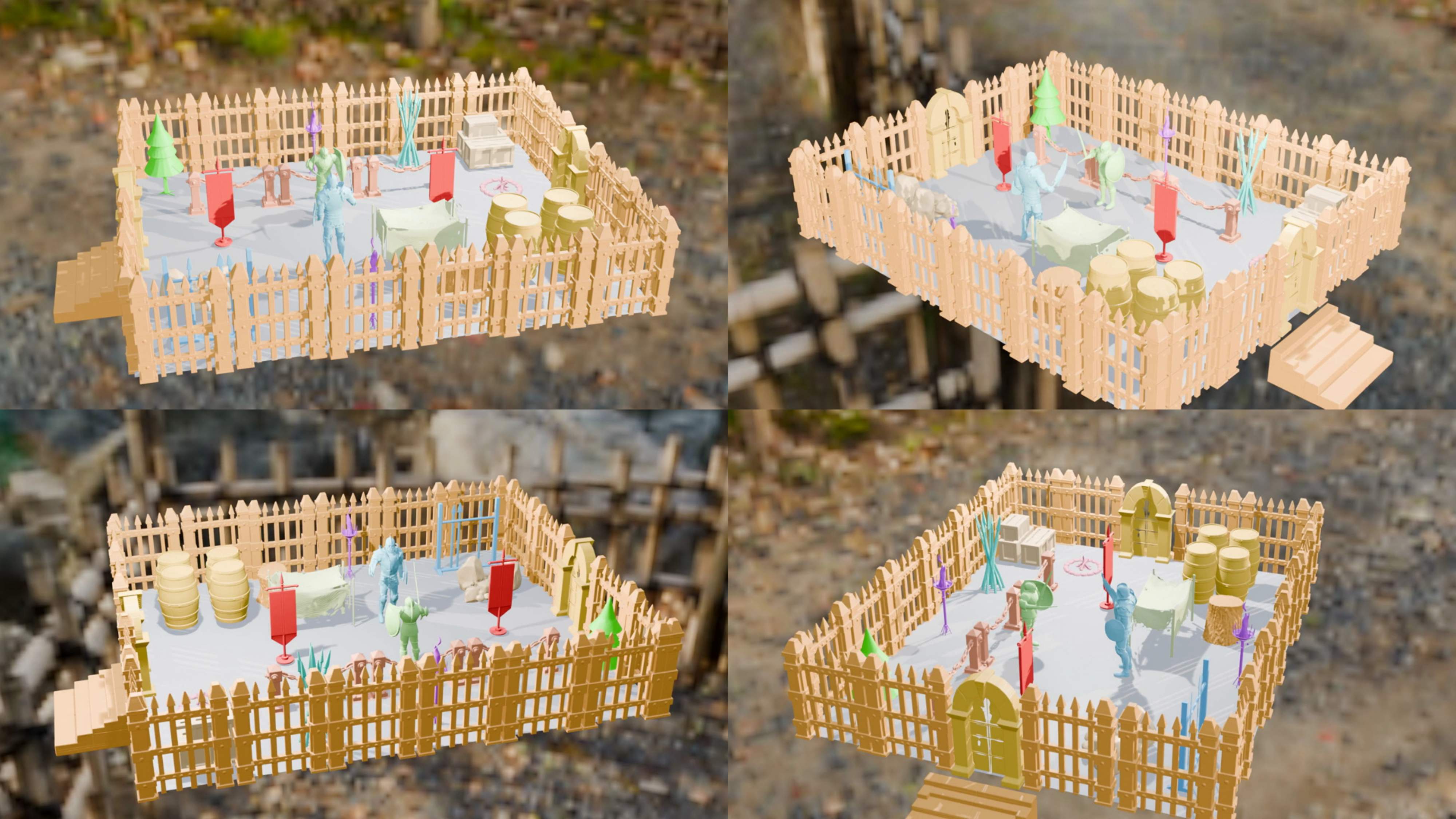}
\caption{Four views of the manually assembled teaser arena. The layout, repeated placement of objects, display colors, lighting, and background are Blender presentation choices; Block3D generates the component meshes rather than the complete arena composition.}
\label{fig:teaser_views}
\includegraphics[width=0.90\textwidth]{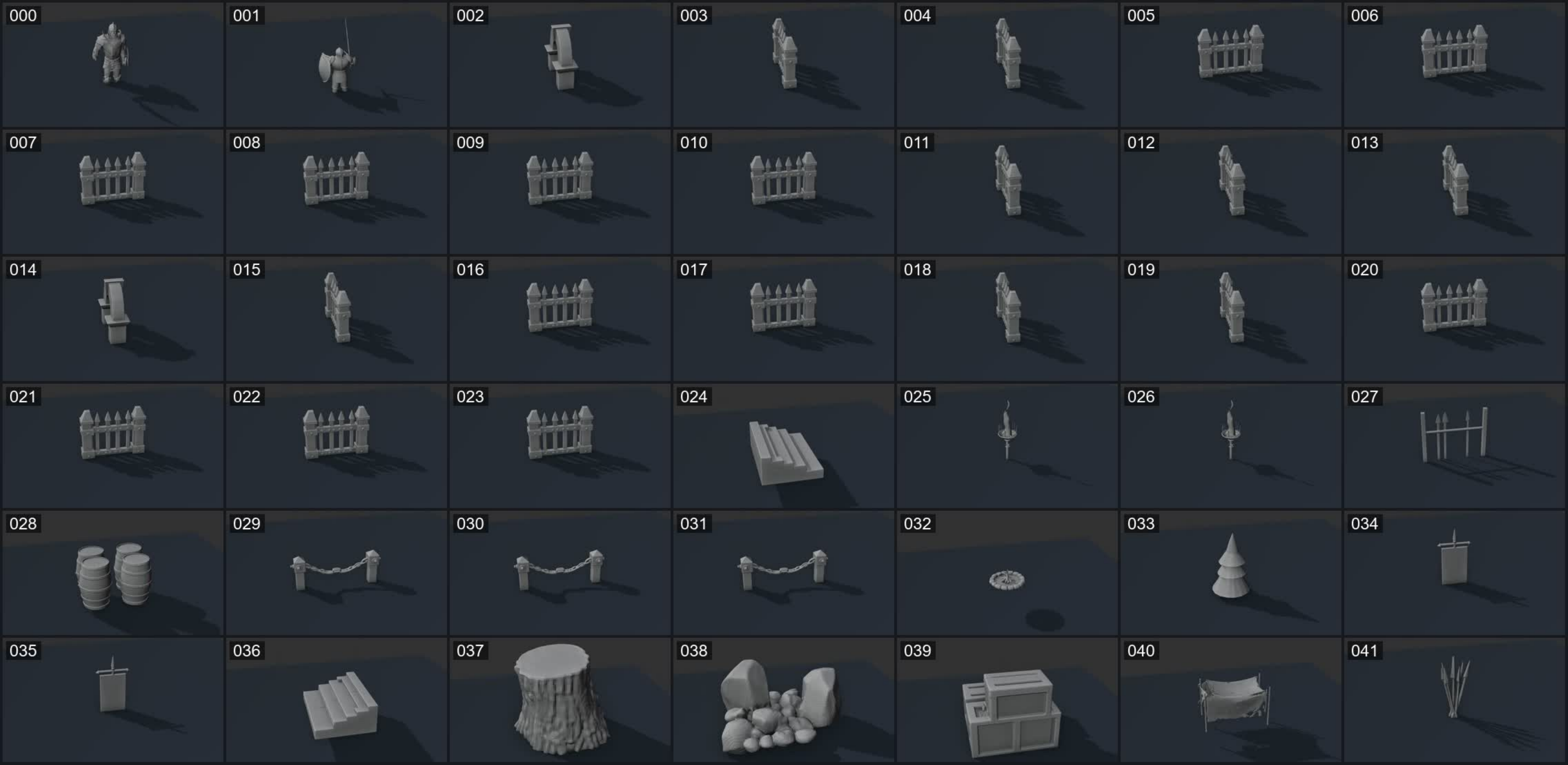}
\caption{Isolated mesh-object inventory for the teaser scene. Labels 000--041 match the media manifest and provide a stable visual index for the individual turntable files.}
\label{fig:isolated_assets}
\end{figure}

\begin{figure}[p]
\centering
\includegraphics[width=0.90\textwidth]{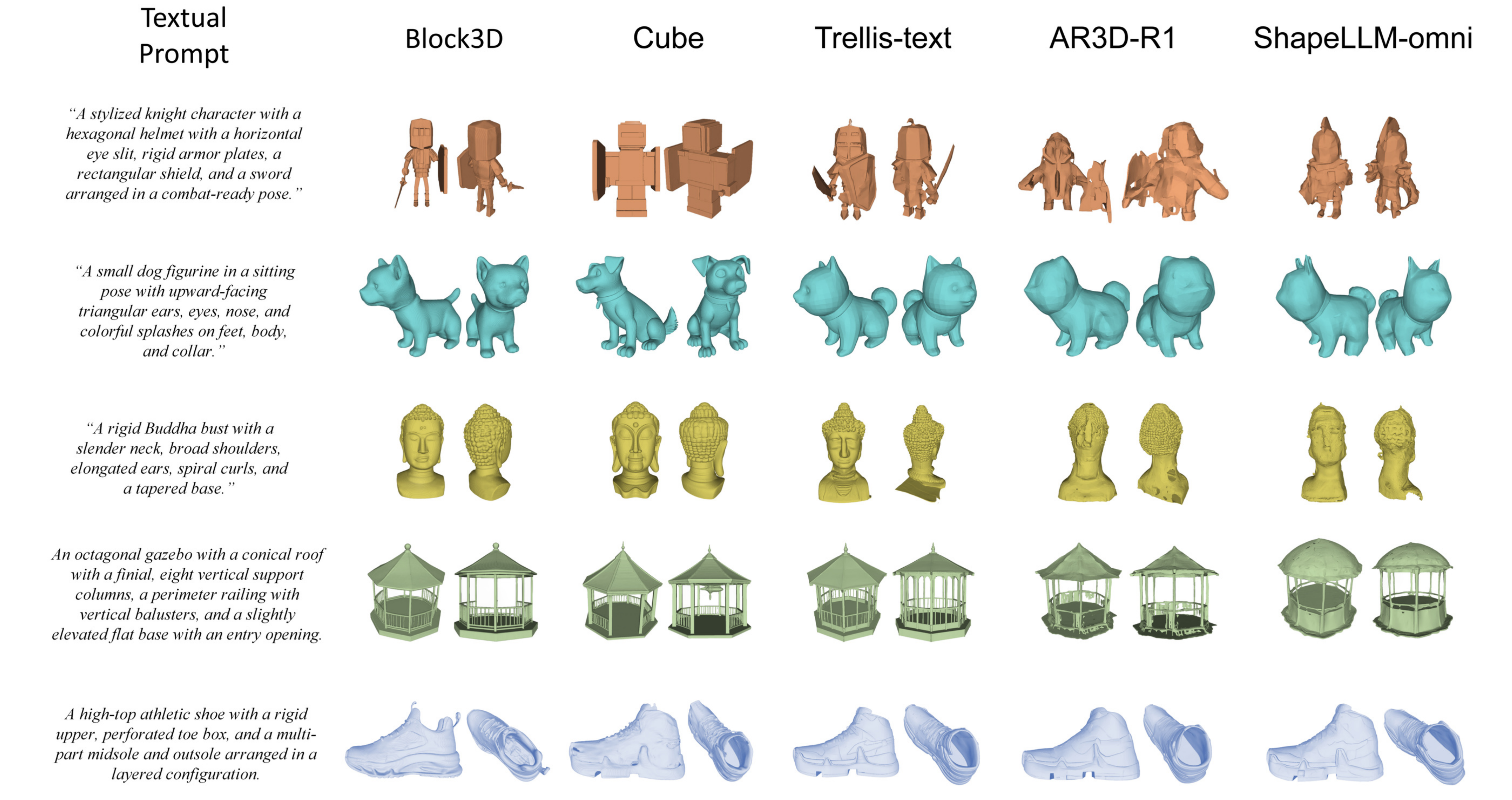}
\caption{Full-prompt version of the qualitative comparison in the main paper. Columns show Block3D, Cube, TRELLIS-text, AR3D-R1, and ShapeLLM-Omni; each result is rendered from front and back views.}
\label{fig:full_qualitative}
\end{figure}

\begin{figure}[p]
\centering
\includegraphics[width=0.94\textwidth]{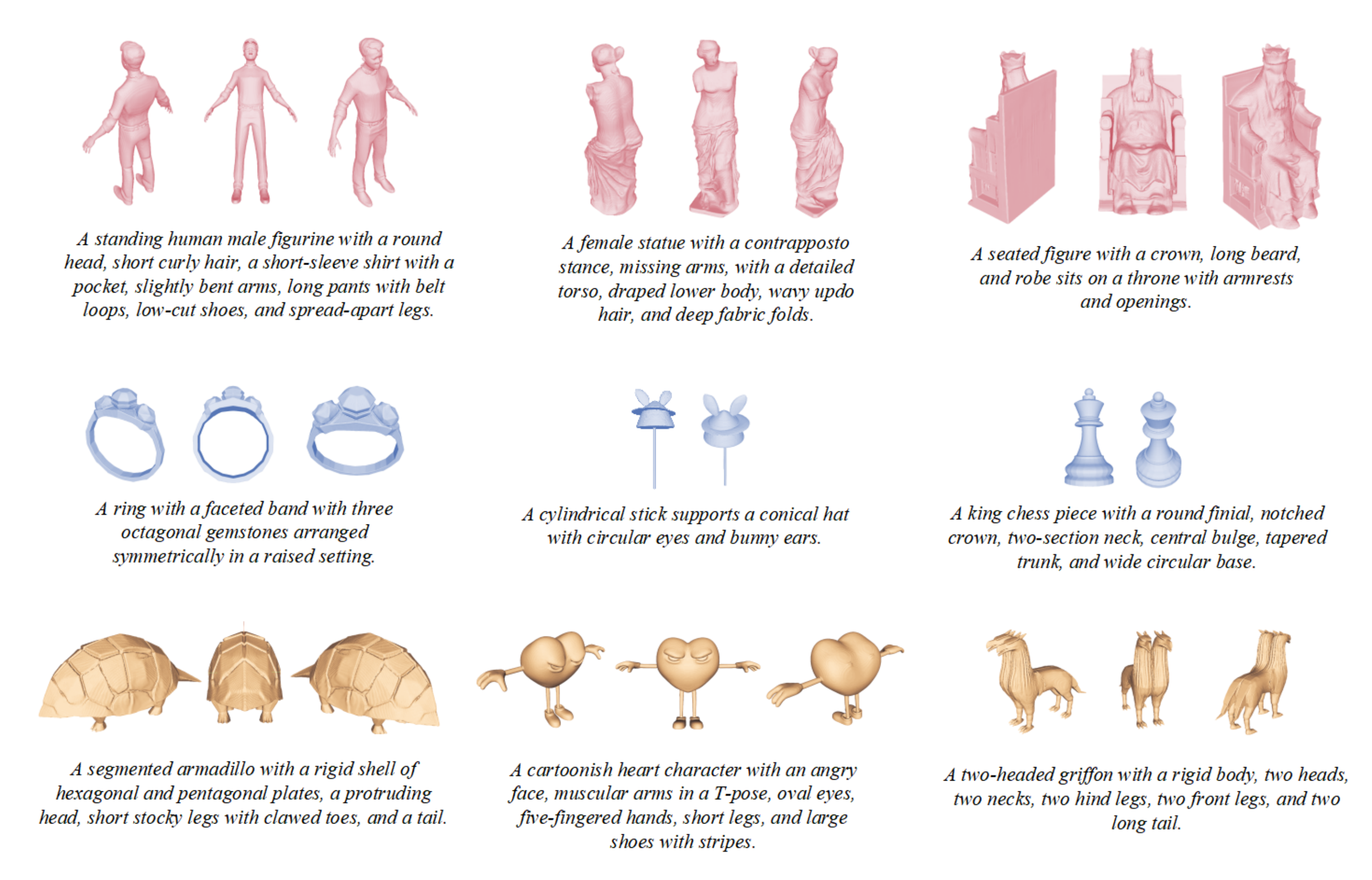}
\caption{Additional Block3D text-to-shape results. Each group contains multiple viewpoints of a single generated mesh, with its complete text prompt shown directly below.}
\label{fig:additional_demos}
\end{figure}

This section documents all qualitative figures before listing their exact prompts. Figures~\ref{fig:teaser_views} and~\ref{fig:isolated_assets} separate the teaser presentation from its generated components, Figure~\ref{fig:full_qualitative} restores the complete prompts used in the main comparison, and Figure~\ref{fig:additional_demos} broadens the visual coverage with additional Block3D outputs.

\subsection{Teaser Prompts and Isolated Assets}
\label{sec:teaser_assets}

The teaser in the main paper is a Blender scene assembled from Block3D mesh outputs. It is not a direct multi-object or scene-generation output. Generated mesh objects were imported into Blender~\cite{blender}, arranged into the arena, assigned display materials, and rendered with a shared scene, camera, and lighting setup.

Each unique component is generated once with the deterministic Block3D configuration in Section~\ref{sec:complete_sampler}. Blender duplicates, transforms, colors, and arranges the meshes but does not modify their geometry.

The assembled arena therefore demonstrates the composability of independently generated assets. It does not claim that Block3D generates a complete scene layout or a jointly conditioned multi-object scene.

\paragraph{Component Prompts.}
The identifiers below match Figure~\ref{fig:isolated_assets}. Each prompt generates one object. Ranges denote repeated scene instances that reuse the same generated mesh; multiplicity and placement are determined only during scene assembly.
\begin{itemize}
\item \textbf{000 (knight).} ``A stylized armored knight with a helmet, plate armor, shield, and sword.''
\item \textbf{001 (dog).} ``A small dog companion with upright ears.''
\item \textbf{002/014 (throne).} ``A high-backed wooden throne with a stepped base.''
\item \textbf{003--023 (fence).} ``A short wooden palisade fence with pointed stakes.''
\item \textbf{024/036 (stairs).} ``A broad three-step stone stair.''
\item \textbf{025/026 (torch).} ``A medieval standing torch on a narrow pole.''
\item \textbf{027 (rack).} ``A simple rack holding upright spears.''
\item \textbf{028 (barrels).} ``Two wooden barrels placed side by side.''
\item \textbf{029--031 (barrier).} ``Two short posts connected by a heavy chain.''
\item \textbf{032 (marker).} ``A circular stone floor marker.''
\item \textbf{033 (tree).} ``A stylized conical pine tree.''
\item \textbf{034/035 (banner).} ``A rectangular hanging arena banner.''
\item \textbf{037 (stump).} ``A cut tree stump with visible roots.''
\item \textbf{038 (rocks).} ``A clustered pile of rocks.''
\item \textbf{039 (chest).} ``A reinforced wooden storage chest.''
\item \textbf{040 (hammock).} ``A cloth hammock suspended between two supports.''
\item \textbf{041 (spears).} ``A tied bundle of long spears.''
\end{itemize}

\subsection{Figure 4 Prompts and Results}
\label{sec:figure4_prompts}

The complete prompts corresponding to the five rows of Figure~4 in the main paper are:
\begin{enumerate}
\item ``A stylized knight character with a hexagonal helmet with a horizontal eye slit, rigid armor plates, a rectangular shield, and a sword arranged in a combat-ready pose.''
\item ``A small dog figurine in a sitting pose with upward-facing triangular ears, eyes, nose, and colorful splashes on feet, body, and collar.''
\item ``A rigid Buddha bust with a slender neck, broad shoulders, elongated ears, spiral curls, and a tapered base.''
\item ``An octagonal gazebo with a conical roof with a finial, eight vertical support columns, a perimeter railing with vertical balusters, and a slightly elevated flat base with an entry opening.''
\item ``A high-top athletic shoe with a rigid upper, perforated toe box, and a multi-part midsole and outsole arranged in a layered configuration.''
\end{enumerate}
Figure~\ref{fig:full_qualitative} reproduces the comparison without truncating these prompts. Each method is shown from the same two semantic viewpoints used in the main figure. The red boxes in the main paper identify missing or distorted local geometry and are presentation annotations rather than additional measurements.

These five prompts constitute a separate qualitative panel chosen to cover an articulated character, an animal, a bust, architecture, and a manufactured object. They are not members of the random100 paired evaluation set and do not contribute to the quantitative averages. The prompt list is fixed and shared across methods, and front and back views use the same camera and normalization for all methods.

\subsection{Additional Text-to-Shape Examples}
\label{sec:additional_demos}

Figure~\ref{fig:additional_demos} presents nine additional Block3D outputs produced with the deterministic configuration in Section~\ref{sec:complete_sampler}. Each group shows two or three rendered viewpoints of one generated mesh, rather than separate generations, and prints the complete generation prompt below the corresponding result. The examples extend the qualitative coverage to human figures and statues, accessories and manufactured objects, and stylized animals and characters. This panel is illustrative and is not included in any reported metric or model-selection decision. Its complementary viewpoints expose the same decoded geometry, including thin structures, bilateral forms, and articulated silhouettes that may be hidden from a single view.

\section{Code and Data Package}
\label{sec:code_data_package}

\subsection{Archive Contents and Implementation Map}
\label{sec:code_index}

The source archive is organized around three reproducibility paths. The \path{block3d/training/} package contains the data loader, clean/corrupted attention construction, sample-level M2T/T2T corruption, no-gradient rollout, residual objective, and distributed trainer. The \path{block3d/inference/} package contains condition preparation for the reported text-only path, the bounded editable sampler, CFG schedule, and prefix-cache execution. Shared masks, reveal schedules, and M2T/T2T update sets are implemented in \path{block3d/model/gpt/block_diffusion_utils.py}; geometry and eight-view CLIPScore are implemented in \path{block3d/evaluation.py}. The archive also includes paper-aligned YAML configurations, command-line scripts, \path{pyproject.toml}, a README with complete commands, the upstream-compatible license, and the ordered \path{random100_evaluation_manifest.jsonl}.

The archive contains source and small metadata only. It does not redistribute the Cube~\cite{bhat2025cube} checkpoint, frozen shape tokenizer, CLIP ViT-L/14~\cite{radford2021learning} weights, TRELLIS-500K~\cite{xiang2025trellis} assets, external-baseline repositories, generated meshes, or trained Block3D checkpoints. The README identifies the required Cube release as \texttt{Roblox/cube3d-v0.5}, specifies the expected local artifact layout, and distinguishes the released initialization from random initialization, which is not the protocol reported in the paper.

\subsection{Training and Inference Entry Points}
\label{sec:execution_index}

The main experiment is specified by \path{block3d/configs/block3d.yaml} and \path{block3d/configs/train_block3d.yaml}; the independently trained M2T-only comparison has a corresponding configuration pair. The script \path{scripts/materialize_ablation_configs.py} writes the independently trained configurations for $B\in\{32,96,128,256\}$ and $T\in\{8,12,20\}$, while the main files provide $B=64,T=4$. Every training configuration retains the 35K-step optimization budget, global batch size 40, seed 42, frozen geometry components, and text-only condition. Distributed training is launched through \texttt{torchrun} and \texttt{block3d.train\_block\_diffusion}; the final 35K-step checkpoint is evaluated without validation-based model selection.

Text-to-mesh inference uses \texttt{python -m block3d.generate}. For the reported protocol, the command omits both the optional bounding-box argument and top-$p$, disables optional PyMeshLab postprocessing, uses $B=64$, $T=4$, $g=3.0$, and writes one native decoded mesh for one prompt. The release deliberately excludes cluster-specific batch scheduling: the ordered evaluation manifest provides the 100 prompts and stable sample IDs, and the same single-prompt entry point is invoked once for each record. This matches the one-generation-per-prompt scope in Section~\ref{sec:statistical_scope}.

\subsection{Data, Evaluation, and Environment Records}
\label{sec:materials_index}

The split script \path{scripts/prepare_block_diffusion_eval_split.py} matches the fixed 100-record manifest to the TRELLIS-500K~\cite{xiang2025trellis} source pool, stops on any missing record or overlap, excludes exactly those 100 assets, and selects the 300K training records with seed 42. It writes the training manifest, the fixed evaluation manifest, the excluded-record list, and a split summary. Third-party mesh targets are referenced by source-relative paths and pair-record keys rather than redistributed.

The script \path{scripts/evaluate_generation.py} consumes per-method \texttt{samples.jsonl} files and writes per-sample JSONL/CSV records plus an aggregate \texttt{summary.json}. Its paper command uses 8,192 surface samples, base seed 0 with the stable record-index offset described in Section~\ref{sec:geometry_metrics}, F@1\% from the stored target bounding-box diagonal, and eight PyTorch3D renders for CLIPScore. Invalid or missing outputs are recorded rather than regenerated. Latency values use the boundary in Section~\ref{sec:latency_protocol}; \path{block3d/benchmarking.py} computes the reported mean, median, P90, and population standard deviation from the ordered records.

Package installation is defined by \texttt{pyproject.toml} for Python 3.10 and PyTorch 2.2.2 or later with CUDA. PyTorch3D is required for the paper-aligned CLIP renderer; Blender and PyMeshLab are optional mesh-import/postprocessing dependencies and do not change the text-only generator. Each training launch stores its resolved model/training configuration, command, random seed, git revision, platform, CPU and memory metadata, Python/PyTorch/CUDA versions, and GPU inventory. These records keep machine-dependent reproduction details separate from the fixed paper protocol stated in Sections~B and~C.

\end{document}